%% file: paper_arxiv.tex
\documentclass{article}

\usepackage[preprint]{vrlbench_preprint}
\makeatletter
\renewcommand{\@noticestring}{Preprint.}
\makeatother
\usepackage[utf8]{inputenc}
\usepackage[T1]{fontenc}
\PassOptionsToPackage{hyphens}{url}
\usepackage[hidelinks]{hyperref}
\usepackage{url}
\usepackage{booktabs}
\usepackage{longtable}
\usepackage{amsmath,amssymb,mathtools}
\usepackage{microtype}
\usepackage[dvipsnames]{xcolor}
\usepackage{graphicx}
\usepackage{array}
\usepackage{multirow}
\usepackage{makecell}
\usepackage{colortbl}
\usepackage{xspace}
\usepackage{float}
\usepackage{placeins}
\usepackage{needspace}
\usepackage{algorithm}
\usepackage{algpseudocode}
\usepackage[capitalize,noabbrev]{cleveref}

\newcommand{\Reflexion}{\textsc{Reflexion}\xspace}
\newcommand{\DCCu}{\textsc{DC-Cu}\xspace}
\newcommand{\VEX}{VEX\textsuperscript{2}\xspace}
\newcommand{\VRLBench}{\textsc{VRL-Bench}\xspace}

\graphicspath{{figures/}{./}}
\title{VRL-Bench: Benchmarking agents on computer\\
control tasks under finite trial budgets}
\input{authors.tex}
\date{}

\hypersetup{
  hypertexnames=false,
  pdftitle={VRL-Bench: Benchmarking agents on computer control tasks under finite trial budgets},
  pdfsubject={Fair evaluation of trial-and-error learning under finite trial budgets},
  pdfkeywords={language agents, trial-and-error learning, verbal reinforcement learning, exploration--exploitation, finite trial budgets, computer control}
}
\begin{document}

\maketitle

\begin{abstract}
\input{sections/abstract.tex}
\end{abstract}

\input{sections/introduction.tex}
\input{sections/evaluation.tex}
\input{sections/results.tex}
\input{sections/scheduler.tex}
\input{sections/related_work_discussion.tex}

\section*{Acknowledgments}
AI tools were used for language polishing and code development.

\appendix
\raggedbottom
\input{appendices/additional_analyses.tex}
\input{appendices/evaluation_results.tex}
\input{appendices/replay.tex}
\input{appendices/scheduler.tex}
\input{appendices/reproducibility.tex}

\end{document}

%% file: authors.tex
\newcommand{\vrlAuthorBlock}{%
\parbox[t]{\dimexpr\textwidth-2\tabcolsep\relax}{%
\centering\bfseries
\mbox{Yu Bai\textsuperscript{1}}\quad \mbox{Yukai Miao\textsuperscript{1}}\quad \mbox{Dawei Wang\textsuperscript{1}}\quad
\mbox{Li Chen\textsuperscript{1}}\quad \mbox{Yanyu Ren\textsuperscript{2}}\quad \mbox{Yuqian Shi\textsuperscript{1}}\quad
\mbox{Dan Li\textsuperscript{2}}\quad \mbox{Ying Xiong\textsuperscript{3}}\quad \mbox{Chengqiu Tan\textsuperscript{3}}\quad
\mbox{Run Zhou\textsuperscript{3}}\quad \mbox{Li Li\textsuperscript{3}}\par
\smallskip\normalfont
\textsuperscript{1} Zhongguancun Laboratory\par
\textsuperscript{2} Tsinghua University\par
\textsuperscript{3} Network Management Center, China Mobile\par
}%
}
\author{\vrlAuthorBlock}
\hypersetup{pdfauthor={Yu Bai; Yukai Miao; Dawei Wang; Li Chen; Yanyu Ren; Yuqian Shi; Dan Li; Ying Xiong; Chengqiu Tan; Run Zhou; Li Li}}

%% file: sections/abstract.tex
Learning from trial and error is a promising way to improve language agents on
complex tasks such as computer control.  \Reflexion introduced verbal
reinforcement learning, which turns failed trials into text that guides later
attempts without updating model parameters.  We introduce \VRLBench, a harness
for fair evaluation of trial-and-error learning under finite trial budgets.
Across three models on MiniWoB and WebShop, we evaluate updates from several
prominent verbal-memory methods spanning \Reflexion and later work: each improves
observed success over memory-free retry in some settings but reduces it in others.
Replay experiments show that using reflection can reduce success rates, revealing
a trade-off between exploiting experience and continued exploration.  We propose
\VEX{}, a verbal exploration--exploitation scheduler that uses a language model
to jointly select policies and allocate the remaining trial budget.  \VEX{} is the
only evaluated update to achieve positive observed success-rate gains over
retry in all six settings.

%% file: sections/introduction.tex
\section{Introduction}
\label{sec:lf-introduction}

Learning from trial and error offers language agents a practical way to improve
on interactive tasks such as computer control.  \Reflexion introduced verbal
reinforcement learning: after a failed complete trial, an agent writes
linguistic feedback, resets the environment, and conditions its next trial on
that text without updating model parameters \citep{shinn2023reflexion}.  The
premise is that this task-specific memory improves later attempts.

A failed trajectory, however, does not determine a unique repair.  A reflection
can focus the next attempt on a useful correction, but an incomplete account can
also anchor the agent to the wrong route while a fresh rollout discovers an
alternative.  Under a finite trial budget, failure-derived memory redirects the
remaining attempts.  Fair evaluation must therefore separate learning across
attempts from the benefit of another chance.

\emph{Memory-free retry} provides the no-update reference: after a reset, the
same fixed-parameter agent receives the task and within-trial observations, but
no text derived from earlier failures.  It need not repeat the same trajectory.
\Reflexion and LATS already include repeated-attempt comparisons, and LATS
reports similar WebShop performance for repeated ReAct and \Reflexion
\citep{shinn2023reflexion,zhou2024lats}.  What remains unclear is how reliably
failure-derived updates improve trial-and-error learning across models and
environments, and when replaying their advice helps rather than hinders.

\begin{figure}[t]
    \centering
    \includegraphics[width=\textwidth]{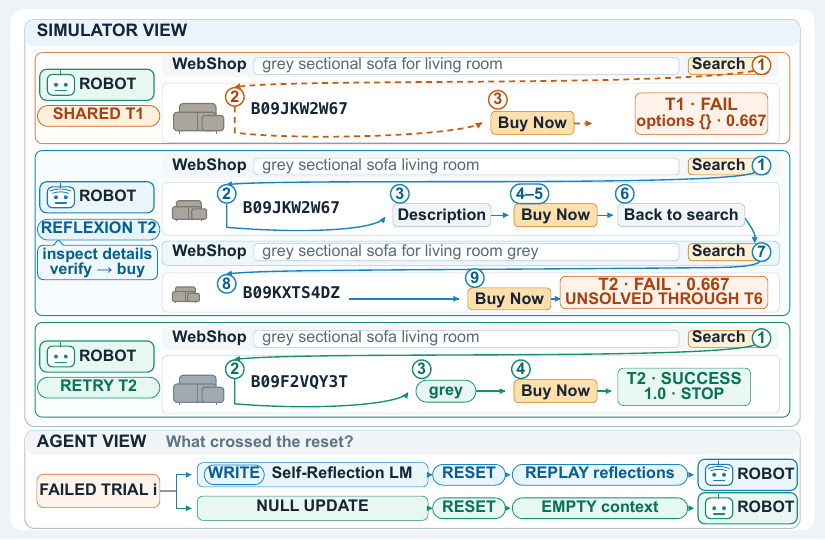}
    \caption{\textbf{Failure memory can redirect trial-and-error away from a
    successful retry.}  In this WebShop pair, both conditions share $T_1$.
    \Reflexion carries a compact failure account forward and remains unsolved
    through $T_6$, whereas retry explores a different product and succeeds at
    $T_2$.  The thought bubble summarizes the stored reflection.}
    \label{fig:lf-reflexion-trial-error}
\end{figure}

We introduce \VRLBench, an evaluation harness for comparing trial-and-error
learning on existing interactive benchmarks, not a new task dataset.  Across
three models on MiniWoB and WebShop \citep{liu2018reinforcement,yao2022webshop},
the comparison reveals a counterintuitive result: every evaluated existing
update helps in some settings and hurts in others.  The harness matches trial
and interaction budgets while preserving each strategy's update, so that the
comparison measures what learning from failure adds to repeated attempts.
\Cref{fig:lf-reflexion-trial-error} illustrates how failure memory can redirect
an agent away from a successful retry.

To understand when memory helps, we revisit \Reflexion's AlfWorld and HotPotQA
experiments with replay-exposure sweeps spanning 3,324 recorded episodes.
Holding the reflection writer fixed, we vary how often its advice reaches the
agent.  Using reflections can reduce success rates, while different replay
schedules recover different failed cases.  These findings reveal an empirical
exploration--exploitation (EE) trade-off
between using failure feedback and exploring alternative continuations.

We propose \VEX{}, a verbal exploration--exploitation scheduler that uses a
language model to decide which policies to exploit or explore within the
remaining trials.  A semantic policy tree, trial allocation, and next-trial
guidance express this joint decision.

Our main contributions are summarized as follows:
\begin{itemize}
    \item We introduce \VRLBench for fair evaluation of trial-and-error learning
    under finite trial budgets and uncover a counterintuitive result: every
    evaluated existing memory update improves on retry in some settings but
    falls below it in others.  The comparison spans three models on MiniWoB
    and WebShop.
    \item We show through controlled experiments varying reflection replay
    frequency that using reflections can reduce success rates, revealing a
    trade-off between exploiting experience and continued exploration.
    \item We propose \VEX{}, which uses a language model as an
    exploration--exploitation scheduler under a finite trial budget.  \VEX{} is the
    only evaluated update to achieve positive observed success-rate gains over
    retry in all six settings.
\end{itemize}

%% file: sections/evaluation.tex
\section{\VRLBench: Fair Evaluation of Trial-and-Error Learning}
\label{sec:lf-problem}
\label{sec:lf-vrlbench}

\VRLBench evaluates how language agents learn across repeated attempts on
existing interactive benchmarks under a finite complete-trial budget.  It
standardizes what counts as an attempt, when the environment resets, when
evaluation stops, and which task feedback is available across attempts, while
preserving each strategy's own cross-trial update and memory representation.

\subsection{Task and Trial Formulation}

For each task, the same language agent, with model parameters unchanged,
receives at most $T$ complete trials and stops at the first success.  We call
$T$ the \emph{complete-trial budget}.  A complete trial begins at the
benchmark-defined reset and ends at terminal success, terminal failure, or
exhaustion of the benchmark-specific interaction limit.  SR@$T$ measures the
fraction of tasks solved within this budget.  Model calls, tokens, and
action--observation transitions are reported separately; they do not count as
additional attempts.

\paragraph{Reset without rollback.}
The harness models computer control in which an agent cannot restore an
intermediate state to test another action branch.  A later attempt starts at
the benchmark's entry state, and any reset-local values must be read afresh.

\paragraph{Fair comparison.}
To isolate what learning across attempts adds, all conditions for a case share
the same realized $T_1$.  If it fails, each strategy applies its own update and
begins $T_2$ from the same external state with the same remaining trial budget
and within-trial interaction limit.  Later trials are fresh and stop at success;
their trajectories may diverge.

\paragraph{Cross-trial information.}
After a failed trial, an update may use the task, public environment rules, the
accumulated action--observation traces and outcomes from earlier counted
failures, and the remaining trial budget.  A strategy may also retain text
generated during those failures, but that text adds no environment observations
beyond the counted trials.  Memory-free retry carries no failure-derived text
across resets.  Appendix~\ref{app:iclr-formal-details} gives the formal
definitions.

\subsection{Benchmarks, Conditions, and Outcomes}

We evaluate MiniWoB and WebShop with DeepSeek-V4-Flash,
GPT-5.4 nano, and GLM-4.7-FlashX
\citep{deepseekai2026v4,openai2026gpt54nano,zai2026glm47}.  MiniWoB has 64 task
families with ten reset-seed sequences each, paired across methods ($N=640$).
Each trial uses a fresh instance of the same family.  Trials permit at most
128 dispatched transitions and a task-specific limit on Actor action blocks,
which may contain several actions
(Appendix~\ref{app:lf-reproducibility-profiles}).  WebShop uses the first 100
official human-test goals, following the evaluation scale of
\Reflexion and ExpeL \citep{shinn2023reflexion,zhao2024expel}; each trial resets the same goal
and permits at most 15 transitions.  All primary
conditions use $T=6$.  Within each model--environment setting, conditions share
the cases, Actor configuration, success rule, reset semantics, and interaction
limits.

\paragraph{Evaluation scale.}
The primary study covers 2,220 paired cases across three models and two
benchmarks.  Evaluating all five conditions requires 16,946 distinct
complete-trial executions with early stopping upon success, counting each shared $T_1$
once.  Appendix~\ref{app:lf-vrlbench-primary} gives the full evaluation matrix.

\paragraph{Compared updates.}
We compare memory-free retry, \Reflexion, DC-Cu$^\ast$, and ACE$^\ast$ across six
settings; \VEX{} is evaluated separately in \cref{sec:lf-dsst}.
\Reflexion appends reflections after eligible failures and passes their ordered
history to the next Actor.  DC-Cu$^\ast$ retains Dynamic Cheatsheet's cumulative
sheet and curator/extractor replacement update \citep{suzgun2025dynamiccheatsheet};
ACE$^\ast$ retains the Reflector and online ADD-only Curator for playbook updates
\citep{zhang2026ace}.  Originally evaluated across queries or tasks, Dynamic
Cheatsheet and ACE require adaptation for fair trial-and-error comparison with
\Reflexion and retry.  We retain their updates but use only within-case failures
under matched trial and interaction budgets, resetting
memory between cases.  We exclude ExpeL because it requires successful
trajectories.  Appendices~\ref{app:lf-source-protocol-details}
and~\ref{app:lf-vrlbench-information-flow} give the original settings,
adaptations, and update inputs.

\paragraph{Outcomes and uncertainty.}
SR@6 directly measures whether a strategy solves the task within six complete
attempts.  We also report recovery after the shared first-trial failure (RR@6)
and censored AvgT@6, which assigns unsolved cases to trial $T+1$.
Retry-relative effects are paired within each model--environment setting, using
64 MiniWoB task-family means and 100 WebShop goals as sampling units.  We report
setting-specific 95\% paired bootstrap intervals.  Metric definitions appear in
Appendix~\ref{app:iclr-formal-details}; bootstrap and testing procedures appear
in Appendix~\ref{app:lf-vrlbench-reevaluation-inference}.

%% file: sections/results.tex
\FloatBarrier
\section{Existing Verbal Updates Do Not Consistently Outperform Retry}
\label{sec:lf-reevaluation}

\begin{figure}[H]
    \centering
    \includegraphics[width=\textwidth]{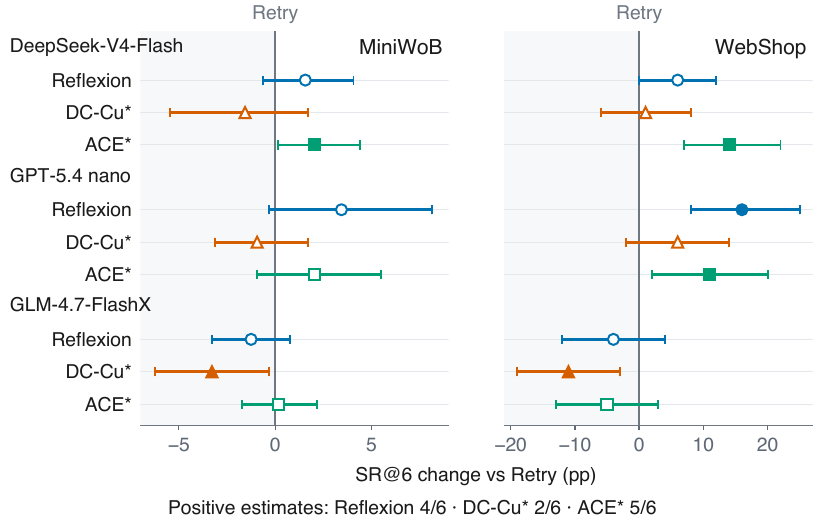}
    \caption{\textbf{None of the three existing verbal updates consistently
    outperforms memory-free retry.}  Panels show MiniWoB and WebShop, with
    aligned rows grouped by model.  Points show SR@6 differences in percentage
    points; bars are setting-specific 95\% paired intervals over 64 MiniWoB
    task-family means or 100 WebShop goals.  Filled markers denote intervals
    excluding zero; open markers include zero.  Horizontal scales differ
    by environment.  Shading marks negative differences; the bottom line counts
    positive point estimates, not intervals excluding zero.}
    \label{fig:lf-reevaluation-paired-effects}
\end{figure}

\emph{Does learning from failure outperform simply trying again under the same
trial budget?}  Under \VRLBench, none of the three existing verbal updates improves SR@6 over
memory-free retry in every setting.  \Cref{tab:lf-primary-sr-main} reports the
absolute outcomes, and \cref{fig:lf-reevaluation-paired-effects} shows the paired
retry-relative differences that establish this result.

The direction of improvement changes with the model, not just the update.
\Reflexion improves SR@6 on both environments with DeepSeek-V4-Flash and
GPT-5.4 nano, but lowers it on both with GLM-4.7-FlashX.  Even ACE$^\ast$,
which is positive in five settings, drops from retry's 50\% to 45\% on
GLM-4.7-FlashX WebShop.  DC-Cu$^\ast$ falls below retry in all three MiniWoB
settings.

The paired intervals support both gains and losses in particular settings.
\Reflexion gains 16 points on GPT-5.4 nano WebShop (95\% CI $[+8,+25]$),
and ACE$^\ast$ has three settings with positive intervals excluding zero.
DC-Cu$^\ast$, by contrast, has negative intervals excluding zero for
GLM-4.7-FlashX in both environments.  Full SR@6, RR@6, AvgT@6 results and
paired intervals appear in Appendix~\ref{app:lf-vrlbench}.

The number of attempts matters as well as eventual success.  No existing
update lowers censored AvgT@6 in all six settings
(\cref{tab:lf-primary-sr-main}).  \Reflexion's trial-efficiency changes follow
the same model-dependent pattern as its success rates: WebShop AvgT@6 falls
from 4.81 to 4.19 with GPT-5.4 nano but rises from 4.39 to 4.46 with
GLM-4.7-FlashX.  The same reflection update can improve both outcomes in one
setting and worsen both in another.

\begin{table}[H]
\centering
\small
\setlength{\tabcolsep}{3.1pt}
\renewcommand{\arraystretch}{1.05}
\begin{tabular*}{\textwidth}{@{\extracolsep{\fill}}lrrrrrr@{}}
\toprule
& \multicolumn{3}{c}{MiniWoB}
& \multicolumn{3}{c}{WebShop} \\
\cmidrule(lr){2-4}\cmidrule(lr){5-7}
Method & \makecell{DeepSeek-V4\\Flash} & \makecell{GPT-5.4\\nano}
& \makecell{GLM-4.7\\FlashX} & \makecell{DeepSeek-V4\\Flash}
& \makecell{GPT-5.4\\nano} & \makecell{GLM-4.7\\FlashX} \\
\midrule
\multicolumn{7}{@{}l}{\textit{SR@6 (\%), higher is better}} \\
Retry (no memory) & 87.3 & 82.5 & 82.5 & 45.0 & 39.0 & 50.0 \\
\Reflexion      & 88.9 & 85.9 & 81.2 & 51.0 & 55.0 & 46.0 \\
\DCCu$^\ast$    & 85.8 & 81.6 & 79.2 & 46.0 & 45.0 & 39.0 \\
ACE$^\ast$      & 89.4 & 84.5 & 82.7 & 59.0 & 50.0 & 45.0 \\
\midrule
\multicolumn{7}{@{}l}{\textit{Censored AvgT@6, lower is better}} \\
Retry (no memory) & 1.99 & 2.46 & 2.40 & 4.41 & 4.81 & 4.39 \\
\Reflexion       & 1.92 & 2.25 & 2.46 & 4.14 & 4.19 & 4.46 \\
\DCCu$^\ast$     & 2.05 & 2.48 & 2.58 & 4.40 & 4.74 & 4.84 \\
ACE$^\ast$       & 1.91 & 2.33 & 2.41 & 3.91 & 4.40 & 4.53 \\
\bottomrule
\end{tabular*}
\caption{\textbf{Existing updates do not consistently improve either success
or trial efficiency over retry.}  SR@6 counts cases solved within six attempts;
AvgT@6 averages the first-success trial, assigning unsolved cases to 7.
Each model uses 640 MiniWoB cases and 100 WebShop goals.  $^\ast$ marks within-task
failure-only adaptations; paired SR@6 intervals appear in
\cref{fig:lf-reevaluation-paired-effects}.}
\label{tab:lf-primary-sr-main}
\end{table}

\FloatBarrier
\section{Replay Exposure Changes the Value of Reflection}
\label{sec:lf-replay-main}

\begin{figure}[!htbp]
    \centering
    \includegraphics[width=\textwidth]{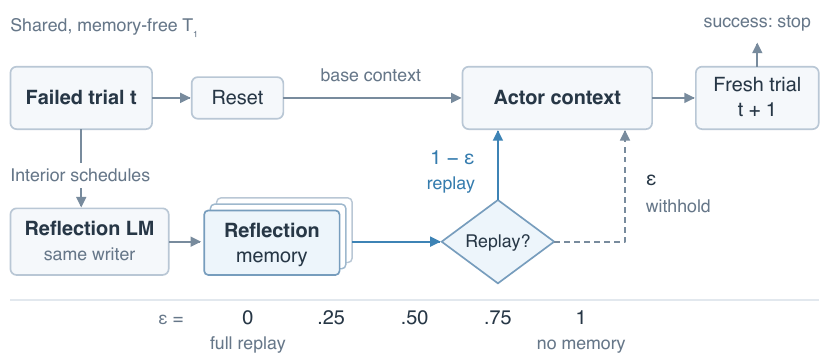}
    \caption{\textbf{Controlling reflection replay separates writing memory
    from using it.}  In the interior conditions, failures update reflection
    memory; before each later trial, a gate exposes it with probability
    $1-\varepsilon$ or withholds it with probability $\varepsilon$, without
    clearing it.  Actor and writer configurations, shared $T_1$, and trial
    limits stay fixed.  Endpoints reuse full-replay and no-memory runs; the
    no-memory endpoint supplies no reflection to the Actor.}
    \label{fig:iclr-replay-exposure-design}
\end{figure}

\emph{Does reflection's early advantage persist across trials?}
\Reflexion demonstrated learning across repeated attempts on AlfWorld and
HotPotQA \citep{shinn2023reflexion}.  We revisit these settings with a controlled
replay intervention that varies how often the agent sees its reflections after
the same first-trial failure.  This separates having a reflection from using
it to guide each remaining attempt (\cref{fig:iclr-replay-exposure-design}).

\paragraph{Replay intervention.}
We use a GPT-4o-mini-backed Actor on 134 AlfWorld environments and 100 HotPotQA
questions, with budgets of six and five trials respectively.  Within each
setting, conditions share the first trial and the trial limits.  For interior
withholding probabilities $\varepsilon\in\{.25,.50,.75\}$, reflection writing
continues after failure, but the next Actor sees the stored reflections with
probability $1-\varepsilon$.  The endpoints are full replay ($\varepsilon=0$)
and memory-free continuation ($\varepsilon=1$).  The five-value grid contains 3,324
recorded episodes with early stopping upon success (Appendix~\ref{app:lf-replay}).

\begin{figure}[!htbp]
    \centering
    \includegraphics[width=\textwidth]{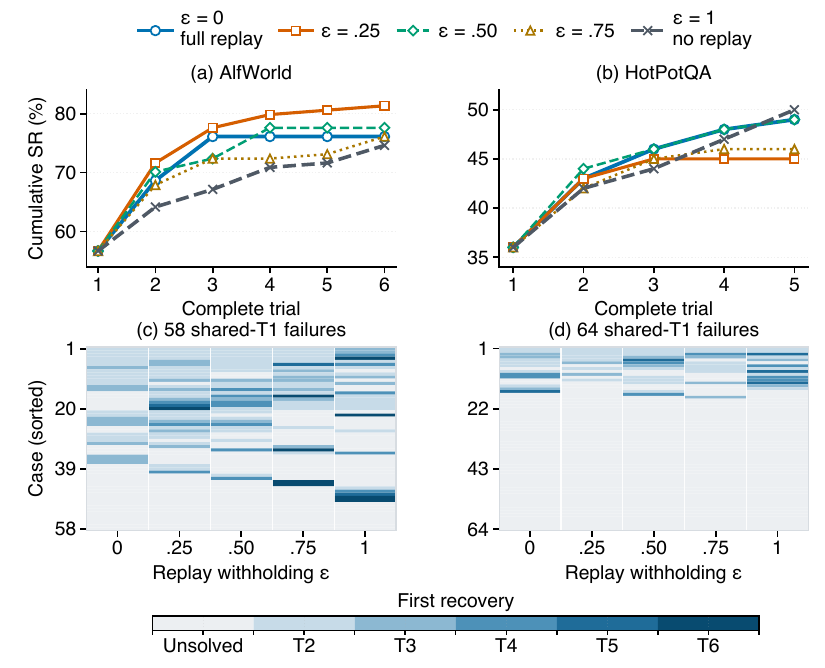}
    \caption{\textbf{Replay's early advantage can disappear, and different
    schedules recover different failures.}  Top: cumulative success under five
    replay schedules on 134 AlfWorld and 100 HotPotQA cases.  HotPotQA full replay
    leads no replay at trials 2--4 but finishes at 49\%, below it at 50\%.
    Bottom: first recovery trial for each shared-$T_1$ failure; gray means
    unsolved.  These are observed paired-cohort patterns; inferential intervals
    appear in Appendix~\ref{app:iclr-additional-analyses}.}
    \label{fig:lf-replay-instance-structure}
\end{figure}

\paragraph{The comparison changes across trials.}
On HotPotQA, full replay leads no replay by 1, 2, and 1 percentage points at
trials 2--4, but falls behind at trial 5: 49\% versus 50\%
(\cref{fig:lf-replay-instance-structure}).  The memory-free condition improves
from 36\% at $T_1$ to 50\%, recovering 14 initially failed questions without
using reflection.  On AlfWorld, full replay reaches 76.1\% at $T_3$ and then
plateaus; $\varepsilon=.25$ continues to 81.3\%, while no replay reaches
74.6\%.  An early advantage for reflection therefore need not persist as
additional attempts become available, and full replay need not yield the
highest observed success rate.

Across the full trial budget, AlfWorld $\varepsilon=.25$ improves mean cumulative
success (AUC) over no replay by 7.1 percentage points, with a selection-adjusted
95\% paired interval of $[+0.9,+13.3]$.  The HotPotQA terminal difference is
$-1.0$ point with a 95\% paired interval of $[-5.0,+3.0]$.
Appendix~\ref{app:iclr-replay-overview} reports the complete paired comparisons.

\paragraph{Recovery paths are complementary.}
The aggregate curves conceal different recovered cases.  Full replay solves
15 AlfWorld cases that no replay misses, while no replay solves 13 that full
replay misses; the corresponding HotPotQA counts are 2 and 3.  Across all five
schedules, the best single schedule recovers 33 of the 58 shared-$T_1$
AlfWorld failures, while their union recovers 49; on HotPotQA the counts are
14/64 and 18/64.  A schedule can therefore succeed where another fails even
when their aggregate success rates are similar.

\Needspace{6\baselineskip}
Together, these results reveal an empirical EE trade-off: using failure feedback
can improve the next attempt, yet different continuations recover failures
that replay misses.  The design question is therefore how to use supported
lessons without committing all remaining trials to one account of failure.
\VEX{} assigns this exploration--exploitation decision to a language model,
conditioned on the failure history and remaining trial budget.  We evaluate
this scheduler next in the six primary settings.

%% file: sections/scheduler.tex
\FloatBarrier
\section{\VEX{}: A Language Model as an Exploration--Exploitation Scheduler}
\label{sec:lf-dsst}

\begin{figure}[!htbp]
    \centering
    \includegraphics[width=\textwidth]{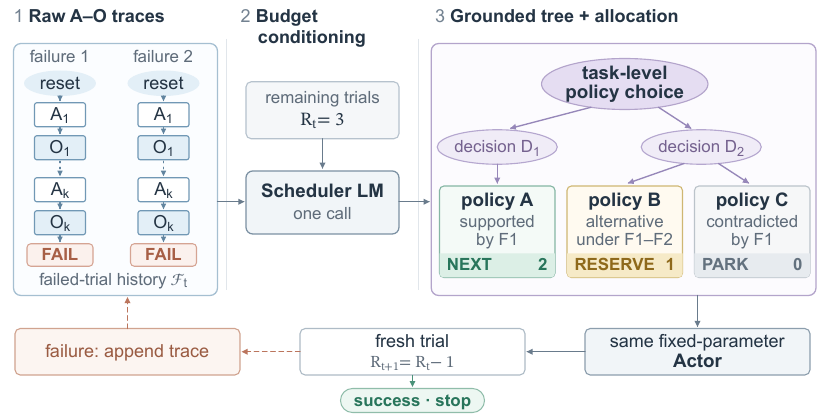}
    \caption{\textbf{\VEX{} uses a language model to schedule exploration and
    exploitation across trials.}  Conditioned on failure history and the remaining
    budget, the model generates a grounded policy tree, an advisory allocation
    requested to sum to $R_t$, and next-trial guidance in one response.
    The complete response passes unchanged to the Actor; only
    eligible failures extend the history.  A realized WebShop response
    appears in Appendix
    Figure~\ref{fig:lf-dsst-information-path-main}.}
    \label{fig:iclr-dsst-compact-method}
\end{figure}

\emph{Can a language model schedule exploration and exploitation across trials
to improve on retry?}  \VEX{} uses a language model as the scheduler: from the
failure history and remaining trial budget, it decides which supported policies
to exploit and which alternatives to explore.  The model expresses this joint
decision in one response containing a semantic policy tree, a trial allocation,
and next-trial guidance (\cref{fig:iclr-dsst-compact-method}).

After a failed trial with $R_t>0$, the language model $\mathcal S_\theta$ maps the task, public
environment rules, ordered failure history, and remaining trial budget to one
response:
\begin{equation}
  \widetilde{\mathcal D}_t
  =\mathcal S_\theta(x,\rho,\mathcal F_t,R_t).
  \label{eq:lf-decision-object}
\end{equation}
The model constructs complete-trial policy hypotheses $\mathcal P_t$ that compete
for the same $R_t$ remaining opportunities.  It assigns an advisory allocation
$\mathbf a_t$, requested to satisfy $\sum_{p\in\mathcal P_t}a_t(p)=R_t$, and
selects a policy $\pi_{t+1}$ to guide the immediate attempt.  Allocating the
budget expresses which alternatives to pursue now, retain for later, or leave
aside, rather than treating every plausible correction as equally actionable.
The tree is a semantic decision space, not an A--O state tree: its branches
describe policies to execute from fresh observations, not intermediate states
that the agent can restore and search.

For example, after a failed foundation purchase, the history shows that the
requested MN4 color was selected successfully even though the purchase failed.
\VEX{} preserves that supported choice while considering a different product
identity.  With five trials left, its response allocates three to a different
sensitive-skin product, two to rechecking the current product's options, and
zero to repeating the unchanged purchase.  The next Actor selects MN4 on a
different product and succeeds at $T_2$
(Appendix~\ref{app:lf-dsst}, \texttt{sample\_0017}).

The Actor receives the complete response unchanged and starts a fresh trial
after environment reset.  The runtime enforces the trial and interaction
limits.  After a failed attempt, any eligible action--observation trace extends
the history.  Whenever a failed attempt leaves budget remaining, the scheduler
is called again, even if the history is unchanged.  Success or budget exhaustion
ends evaluation.  The prompt specification, runtime details, and a
realized WebShop response appear in
Appendix~\ref{app:lf-dsst}.

\subsection{Success and Trial Efficiency}
\label{sec:lf-integrated-results}

Among the four evaluated failure-derived updates, \VEX{} is the only one with a
positive retry-relative SR@6 point estimate in every primary setting
(\cref{tab:lf-dsst-retry-outcomes}).  Gains range from 1.6 to 6.6 percentage
points on MiniWoB and from 8.0 to 17.0 on WebShop.  DeepSeek-V4-Flash and
GPT-5.4 nano improve in both environments with paired intervals above zero;
the two GLM-4.7-FlashX intervals include zero.  Relative to the best result in
\cref{tab:lf-primary-sr-main} for each setting and metric, \VEX{} improves
observed SR@6 by 1.0--8.0 points and lowers censored AvgT@6 in all six settings.

\begin{table}[!htbp]
\centering
\small
\setlength{\tabcolsep}{2.5pt}
\renewcommand{\arraystretch}{1.05}
\begin{tabular*}{\textwidth}{@{\extracolsep{\fill}}lrcrrrr@{}}
\toprule
& \multicolumn{3}{c}{SR@6} & \multicolumn{3}{c}{Censored AvgT@6} \\
\cmidrule(lr){2-4}\cmidrule(lr){5-7}
Model & \VEX{} (\%) & \makecell{$\Delta$ vs Retry (pp)\\{}[95\% paired CI]}
& \makecell{$\Delta$ vs Best\\(pp)} & \VEX{}
& \makecell{$\Delta$ vs\\Retry} & \makecell{$\Delta$ vs\\Best} \\
\midrule
\multicolumn{7}{@{}l}{\textit{MiniWoB}} \\
DeepSeek-V4-Flash & \textbf{93.9} & +6.6 [+2.2, +12.0] & $+4.5^{\mathrm A}$ & \textbf{1.75} & $-0.234$ & $-0.159^{\mathrm A}$ \\
GPT-5.4 nano     & \textbf{87.0} & +4.5 [+0.5, +9.4] & $+1.1^{\mathrm F}$ & \textbf{2.23} & $-0.227$ & $-0.019^{\mathrm F}$ \\
GLM-4.7-FlashX   & \textbf{84.1} & +1.6 [$-1.1$, +4.4] & $+1.4^{\mathrm A}$ & \textbf{2.37} & $-0.028$ & $-0.028^{\mathrm R}$ \\
\midrule
\multicolumn{7}{@{}l}{\textit{WebShop}} \\
DeepSeek-V4-Flash & \textbf{61.0} & +16.0 [+9.0, +23.0] & $+2.0^{\mathrm A}$ & \textbf{3.76} & $-0.650$ & $-0.150^{\mathrm A}$ \\
GPT-5.4 nano     & \textbf{56.0} & +17.0 [+10.0, +25.0] & $+1.0^{\mathrm F}$ & \textbf{4.17} & $-0.640$ & $-0.020^{\mathrm F}$ \\
GLM-4.7-FlashX   & \textbf{58.0} & +8.0 [$-1.0$, +17.0] & $+8.0^{\mathrm R}$ & \textbf{4.08} & $-0.310$ & $-0.310^{\mathrm R}$ \\
\bottomrule
\end{tabular*}
\caption{\textbf{\VEX{} achieves the highest observed success rates and lowest
censored AvgT@6 in all six settings.}
Bold marks the best values across Tables~\ref{tab:lf-primary-sr-main}
and~\ref{tab:lf-dsst-retry-outcomes}.  Deltas subtract Retry or
Table~\ref{tab:lf-primary-sr-main}'s best result for each setting and metric
before rounding.  Superscripts identify the best baseline:
A=ACE$^\ast$, F=\Reflexion, R=Retry.
Best-baseline differences are descriptive; 95\% paired intervals compare
with Retry, using 64 MiniWoB task-family means (640 cases) or 100 WebShop goals.
AvgT@6 assigns unsolved cases to 7.}
\label{tab:lf-dsst-retry-outcomes}
\end{table}

\paragraph{Recovery after the shared failure.}
Paired outcomes distinguish additional recoveries from exchanges of successes
and failures.  On DeepSeek-V4-Flash and GPT-5.4 nano WebShop, \VEX{} solves 16 and
17 goals that retry misses, respectively, while retaining every retry success
in these cohorts.  GLM-4.7-FlashX has a different pattern: \VEX{} solves
15 goals that retry misses but misses 7 that retry solves, yielding the
net gain of 8.  The recovery benefit also appears on MiniWoB: with
DeepSeek-V4-Flash, \VEX{} recovers 118 of the 157 shared first-trial failures,
compared with 76 for retry (Appendix~\ref{app:lf-vrlbench-recovery}).

\paragraph{Success and use of the remaining trials.}
\VEX{} lowers censored AvgT@6 in every setting, primarily by recovering more
failures.  On DeepSeek-V4-Flash WebShop, changed terminal outcomes account for
$0.64$ of the $0.65$ reduction; earlier success among goals solved by both
methods contributes only $0.01$.  On GLM-4.7-FlashX MiniWoB, those common successes are slightly later
under \VEX{}, partially offsetting its additional recoveries.  Thus improved
AvgT@6 need not mean faster completion on already-solvable cases.
Appendix~\ref{app:lf-vrlbench-trial-timing} gives the six-setting decomposition;
mean executed trials, which count unsolved cases as six rather than seven,
also decrease in all six settings.

%% file: sections/related_work_discussion.tex
\par
\FloatBarrier
\Needspace{5\baselineskip}
\section{Related Work}
\label{sec:lf-related-work}

\paragraph{Evaluation methodology for interactive agents.}
Terminal-Bench evaluates terminal agents through containerized environments and
outcome-based tests \citep{merrill2026terminalbench}, while HAL standardizes
evaluation across models, agent scaffolds, and benchmarks \citep{kapoor2026hal}.
AI Agents That Matter motivates strong baselines and cost-aware comparisons,
showing that resampling can match elaborate agents on code-generation tasks
\citep{kapoor2025agentsthatmatter}.  \VRLBench evaluates trial-and-error learning:
whether memory updates improve success and trial efficiency beyond simply trying
again under common trial and interaction budgets.

\paragraph{Learning from verbal feedback and memory.}
Verbal updates operate at different time scales.  Self-Refine and CRITIC revise
outputs using linguistic or tool-grounded feedback
\citep{madaan2023selfrefine,gou2024critic}.  \Reflexion carries failure reflections
across complete trials after reset \citep{shinn2023reflexion}.
MemPrompt, Synapse, ExpeL, Agent Workflow Memory, Dynamic Cheatsheet, and ACE reuse
experience across users, queries, or tasks
\citep{madaan2022memprompt,zheng2023synapse,zhao2024expel,wang2024awm,
suzgun2025dynamiccheatsheet,zhang2026ace}.  We adapt Dynamic Cheatsheet and ACE
to \Reflexion's cross-trial setting, using failures within the current task or
task-family case for fair comparison with \Reflexion and retry.

\paragraph{Search and exploration.}
Search methods differ in where they evaluate alternatives.
Agent-Pro evolves textual policies through development games, with search and
verification where enabled \citep{zhang2024agentpro}.  Tree of Thoughts, RAP,
and LATS generate and evaluate branches during inference
\citep{yao2024tree,hao2023reasoning,zhou2024lats}.  Unlike LATS's rollback search
over action--observation states, \VEX{} uses a language model to construct
semantic policies and allocate fresh complete trials among them.  Its
exploration--exploitation decision concerns which policy to pursue with the
remaining trial budget \citep{sutton2018reinforcement,auer2002finite}.

\section{Discussion and Limitations}
\label{sec:lf-discussion}

The replay sweeps point to a cross-trial decision: which failure-supported
policy to pursue and which alternatives to keep available.  Early replay
benefits need not persist, and different schedules recover complementary
failures.  Action-sequence novelty alone does not capture this trade-off.
In the interior schedules, exposure-weighted estimates show fewer exact repeats
with replay than without it (10.9\% versus 37.6\% on AlfWorld; 4.9\% versus
20.0\% on HotPotQA), while most changed sequences still fail
(Appendix~\ref{app:lf-replay-behavior}).  \VEX{} makes the language model responsible
for this cross-trial trade-off: it chooses which policies to exploit or explore
under the remaining trial budget, rather than seeking a different action
sequence for its own sake.

The evaluation covers three models, MiniWoB and WebShop, six-trial
budgets, and visible environment feedback.  \VRLBench matches trial and
interaction opportunities, not model calls or token use.  The replay sweeps use
separate agent configurations and task sets and establish schedule sensitivity,
not a universal schedule.  Other models, budgets, reset semantics, irreversible
actions, and hidden feedback remain open.

\section{Conclusion}
\label{sec:lf-conclusion}

Our evaluation exposes a counterintuitive gap in memory-based trial-and-error
learning: across three models and two interactive benchmarks, the evaluated
existing updates, from \Reflexion to more recent memory methods, do not
consistently outperform memory-free retry in success or trial efficiency.
This raises a fundamental question: how can agents reliably benefit from failure
experience beyond simply trying again under finite trial budgets?
Replay-exposure sweeps reveal an empirical exploration--exploitation trade-off:
reflection's early advantage can disappear, and different schedules recover
complementary failures.  \VEX{} offers a promising direction by using a language
model to jointly select policies and allocate remaining trials.  It is the only
evaluated update with positive retry-relative SR@6 point estimates and lower
observed censored AvgT@6 in all six settings.  \VRLBench provides a common
framework for investigating this open question through fair evaluation of
trial-and-error learning.

%% file: appendices/additional_analyses.tex
\section{Formal Details of the Cross-Trial State}
\label{app:iclr-formal-details}

Let $h_{t,k}^{(m)}$ be the Actor-visible within-trial history before interaction
step $k$.  With model parameters $\theta$ fixed, actions and observations obey
\begin{align}
  a_{t,k}^{(m)}&\sim\pi_\theta(\cdot\mid x,\rho,C_t^{(m)},h_{t,k}^{(m)}),\\
  o_{t,k+1}^{(m)}&\sim P_{x,\rho}(\cdot\mid h_{t,k}^{(m)},a_{t,k}^{(m)}).
  \label{eq:lf-step-process}
\end{align}
A complete trial yields a trajectory $\tau_t^{(m)}$, terminal outcome
$Y_t^{(m)}$, and public close reason $\kappa_t^{(m)}$.  Let
$T_\star^{(m)}=\min\{t\leq T:Y_t^{(m)}=1\}$, with $T_\star^{(m)}=T+1$ when no
trial succeeds.  The finite-budget objective is
\begin{equation}
  J_T^{(m)}=\Pr\!\left(T_\star^{(m)}\le T\right),
  \label{eq:lf-fixed-budget-objective}
\end{equation}
estimated by SR@$T$.  The benchmark-visible record of trial $t$ is
\begin{equation}
e_t^{(m)}=(o_{t,1}^{(m)},\tau_t^{(m)},Y_t^{(m)},\kappa_t^{(m)}),
\label{eq:lf-benchmark-trial-record}
\end{equation}
and its eligibility to extend cross-trial state is
\begin{equation}
\eta_t^{(m)}=\mathbb I[e_t^{(m)}\text{ is a valid public policy record and }
|\tau_t^{(m)}|\ge 1].
\label{eq:lf-policy-evidence-eligibility}
\end{equation}
Every complete trial consumes one opportunity; $\eta_t^{(m)}$ controls only
whether it may enlarge the cross-trial state.  After trial $t$, the ordered
eligible failure records and any admissible method-internal records are
\begin{equation}
\begin{aligned}
\mathcal F_t^{(m)}&=\operatorname{ord}\{(j,e_j^{(m)}):j\le t,
Y_j^{(m)}=0,\eta_j^{(m)}=1\},\\
\mathcal Q_t^{(m)}&=\operatorname{ord}\{(j,q_j^{(m)}):j\le t,
Y_j^{(m)}=0,\eta_j^{(m)}=1\}.
\end{aligned}
\label{eq:lf-failure-records}
\end{equation}
$R_t=T-t$ counts the complete trials remaining after trial $t$, and the common
external state is
\begin{equation}
  z_t^{(m)}=(x,\rho,\mathcal F_t^{(m)},R_t),
  \label{eq:lf-failure-forest}
\end{equation}
where $x$ is the task or stable task-family scope and $\rho$ contains public
environment rules.
$\mathcal Q_t^{(m)}$ contains only endogenous text generated inside the same
eligible failures and cannot supply a new external observation.  The common
state in \cref{eq:lf-failure-forest} therefore fixes the external information
boundary while allowing method-faithful internal computation.

Let $M_t^{(m)}$ denote method $m$'s persistent cross-trial object.  Each method
may project the common external state and its admissible endogenous text before
forming the next-trial context:
\begin{equation}
  \bigl(M_{t+1}^{(m)},C_{t+1}^{(m)}\bigr)
  =U_m\!\left(M_t^{(m)};V_m(z_t^{(m)}),Q_m(\mathcal Q_t^{(m)})\right).
  \label{eq:lf-method-specific-update}
\end{equation}
$V_m$ and $Q_m$ are method-specific projections; $U_m$ applies the update and
forms the context seen by the next Actor.  This notation defines a common
information boundary, not a common prompt, serialization, or memory format.

The remaining trial budget $R_t$ is a state coordinate, not merely a loop
counter.  If two runs have identical update inputs but different remaining
trial budgets, the update cannot condition its next-trial policy on that
budget difference.  This representational fact does not show that exposing
$R_t$ by itself improves outcomes.

\paragraph{Outcome metrics.}
For $N$ cases under a given method, let
$T_i^\star\in\{1,\ldots,T,T+1\}$ be the first successful trial, with
$T_i^\star=T+1$ when case $i$ remains unsolved, and let
$Y_i(t)=\mathbb I[T_i^\star\leq t]$.  We report
\begin{align}
\mathrm{SR@}T
  &=\frac{1}{N}\sum_iY_i(T),
&
\mathrm{RR@}T
  &=\frac{\sum_i\mathbb I[2\leq T_i^\star\leq T]}
          {\sum_i\mathbb I[T_i^\star>1]},\nonumber\\
\mathrm{AvgT@}T
  &=\frac{1}{N}\sum_iT_i^\star.
\label{eq:lf-vrlbench-metrics}
\end{align}
RR@$T$ measures recovery among cases whose shared first trial failed; AvgT@$T$
censors unsolved cases at $T+1$.

\input{appendices/comparisons.tex}

\section{Additional Analyses of Exploration--Exploitation}
\label{app:iclr-additional-analyses}

This section reports process and diagnostic views of replay exposure and
\VEX{}.  Their settings and numerical results are
expanded in the subsequent appendices.

\FloatBarrier
\subsection{Replay-exposure surface and case structure}
\label{app:iclr-replay-overview}
\FloatBarrier

\begin{figure}[H]
\centering
\includegraphics[width=\textwidth]{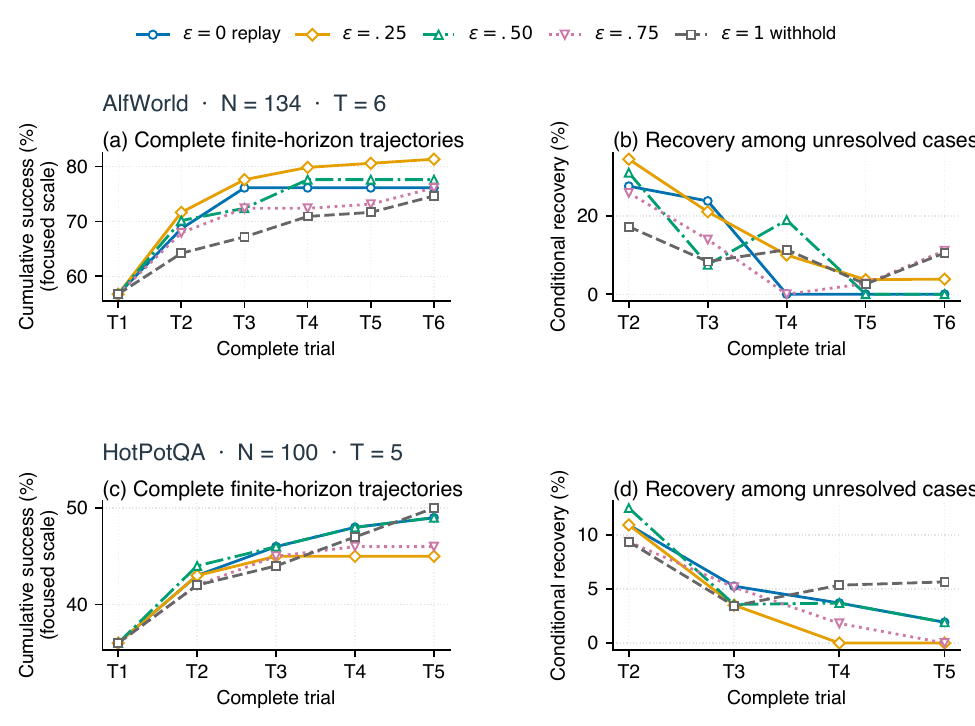}
\caption{\textbf{Endpoint SR@$T$ varies non-monotonically with replay exposure in
the two canonical settings.}  With a GPT-4o-mini-backed Actor, full replay ends
at 49\% on HotPotQA, below full withholding at 50\%; on AlfWorld, the interior
$\varepsilon=.25$ schedule ends highest at 81.3\%, versus 76.1\% for full
replay and 74.6\% for full withholding.  The grid maps schedule sensitivity in
these settings rather than identifying an optimal schedule or general law.}
\label{fig:lf-replay-dynamics}
\end{figure}

The left and right panels distinguish cumulative success from recovery among
cases still unsolved.  Conditional recovery at trial $t$ is
\[
h_t=\frac{\#\{\text{instances first solved at }t\}}
          {\#\{\text{instances unsolved after }t-1\}},\qquad t\geq2.
\]
The changing denominator makes $h_t$ an at-risk conversion rate, not an
unconditional per-trial success rate.  Fixed-budget AUC averages cumulative
success over trials $1$ through $T$; the paired table below reports its
difference alongside terminal SR@$T$.

\begin{figure}[H]
\centering
\includegraphics[width=0.94\textwidth]{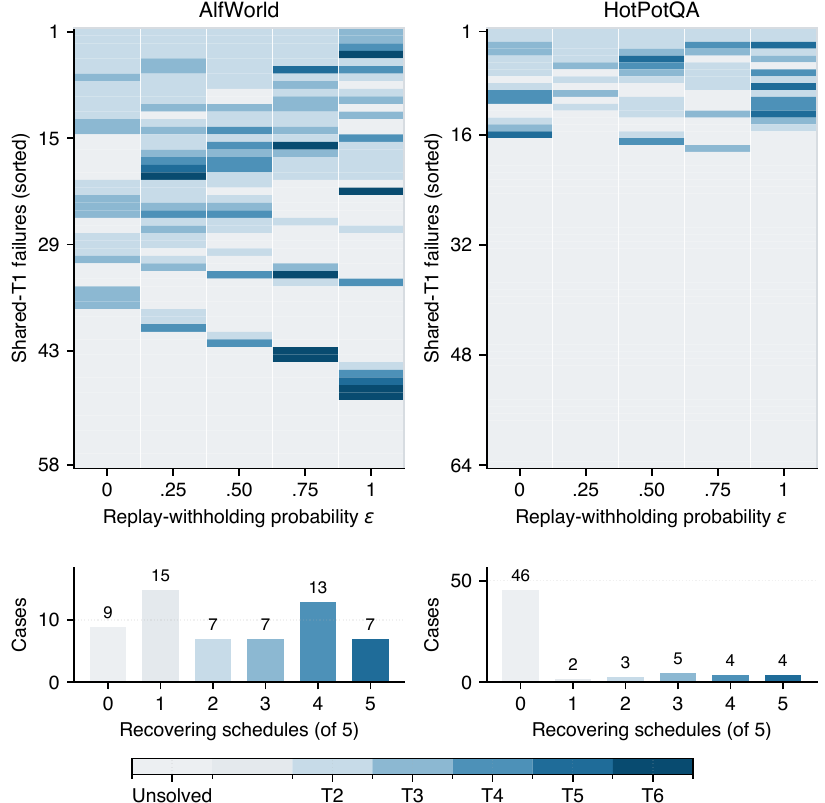}
\caption{\textbf{Replay schedules recover complementary subsets of first-trial
failures.}  The heatmaps show first recovery trials for the 58 AlfWorld and
64 HotPotQA shared-$T_1$ failures.  The lower panels count how many of the five
fixed schedules recover each case.  The union recovers 49 and 18 cases,
respectively, versus 33 and 14 for the best single schedule.}
\label{fig:lf-replay-support}
\end{figure}

\begin{table}[H]
\centering
\small
\setlength{\tabcolsep}{3.2pt}
\begin{tabular*}{\textwidth}{@{\extracolsep{\fill}}llrr@{}}
\toprule
Setting & Exposure comparison & $\widehat\Delta_{\mathrm{SR@}T}$ [95\% CI] & $\widehat\Delta_{\mathrm{AUC}}$ [95\% CI] \\
\midrule
AlfWorld & Full replay $\rightarrow\varepsilon=.25$ & +5.2 [-3.7, +14.2] & +3.0 [-3.2, +9.2] \\
AlfWorld & Full withholding $\rightarrow\varepsilon=.25$ & +6.7 [-2.2, +15.7] & +7.1 [+0.9, +13.3] \\
HotPotQA & Full withholding $\rightarrow$ full replay & -1.0 [-5.0, +3.0] & +0.6 [-2.4, +3.8] \\
\bottomrule
\end{tabular*}
\caption{Paired percentage-point endpoint and AUC differences for the replay
grids.  AlfWorld intervals are simultaneous fixed-grid selection-adjusted;
HotPotQA intervals are pointwise.}
\label{tab:lf-replay-inference}
\end{table}

Of the six endpoint and AUC contrasts in \cref{tab:lf-replay-inference}, only
the AlfWorld AUC gain for $\varepsilon=.25$ over no replay has an interval
entirely above zero; that interval includes fixed-grid selection adjustment
(Appendix~\ref{app:lf-replay-grid}).

\begin{figure}[H]
\centering
\includegraphics[width=\textwidth]{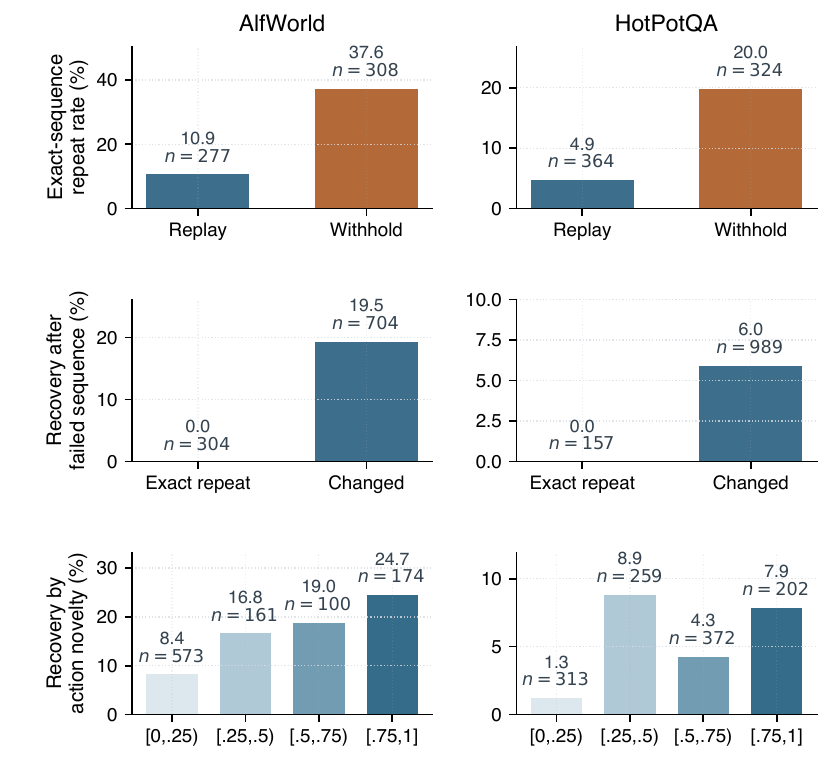}
\caption{\textbf{Every recorded recovery changes the previously failed action
sequence, but change alone is rarely sufficient.}  Exact repeats recover in
0/304 AlfWorld and 0/157 HotPotQA later episodes, versus 137/704 and 59/989
after changed sequences.  Top panels use inverse-probability-standardized
repeat rates over the three interior schedules; the lower two rows pool later
episodes across all five schedules.  $n$ denotes unweighted episode counts.
These associations are descriptive, not causal.}
\label{fig:lf-replay-behavior}
\end{figure}

\FloatBarrier
\Needspace{0.80\textheight}
\subsection[\VEX{} process path and post-T1 recovery]{\VEX{} process path and post-$T_1$ recovery}

\begin{figure}[H]
\centering
\includegraphics[width=0.98\textwidth]{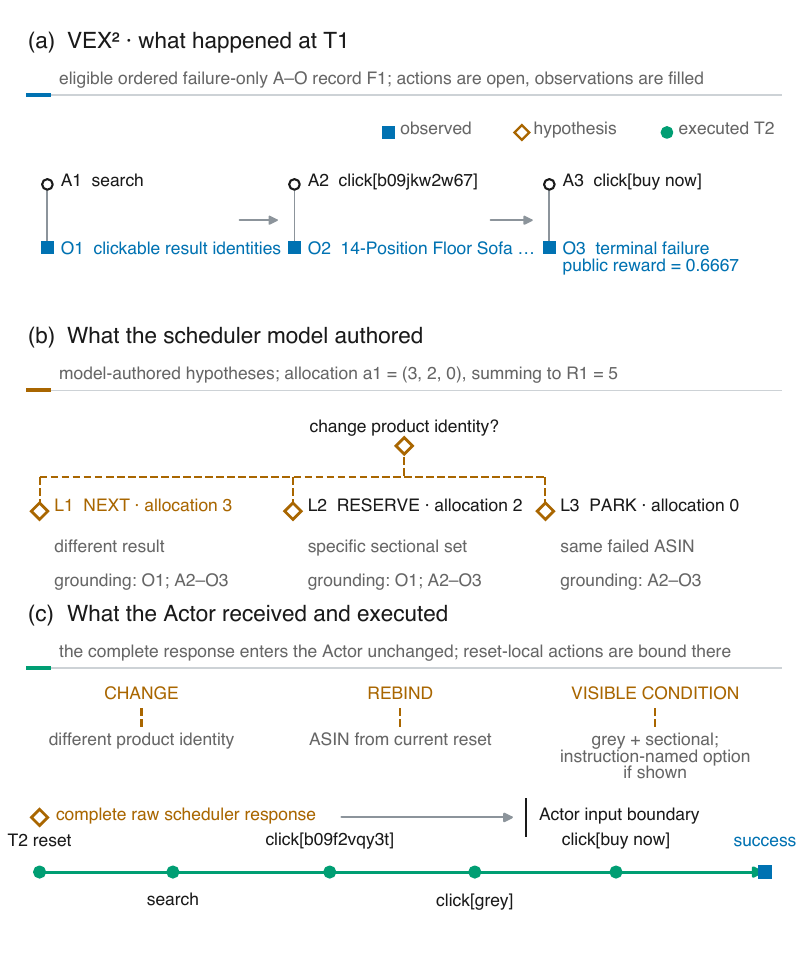}
\caption{\textbf{A realized \VEX{} response jointly schedules the next trial.}
In GPT-5.4-nano WebShop \texttt{sample\_0009}, blue marks the executed $T_1$
history, orange is the single raw response containing policy hypotheses and
an allocation $(3,2,0)$ summing to $R_1=5$, and green is the executed $T_2$
continuation.  The policy headings in (c) summarize the response for readers;
the Actor receives the complete response unchanged.}
\label{fig:lf-dsst-information-path-main}
\end{figure}

\begin{figure}[H]
\centering
\includegraphics[width=\textwidth]{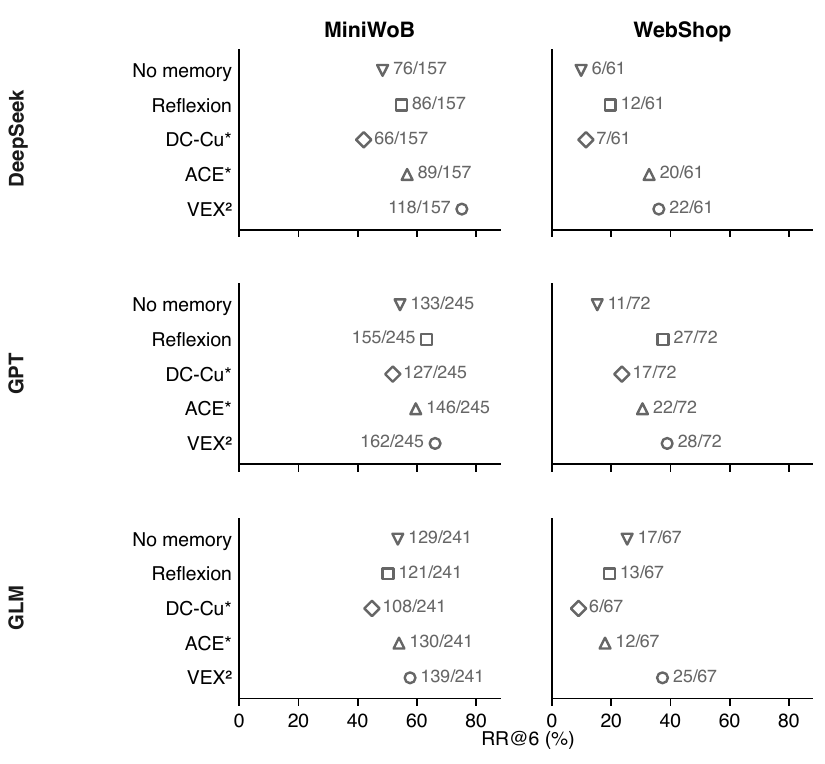}
\caption{\textbf{Among the five evaluated conditions, \VEX{} has the highest
observed post-$T_1$ recovery in all six settings.}  Points report RR@6; labels give recovered cases over the exact
shared-$T_1$ failure denominator.  Counts are descriptive; setting-specific
paired intervals carry the inferential claims.}
\label{fig:lf-vrlbench-recovery}
\end{figure}

%% file: appendices/comparisons.tex
\section{Existing Methods Under a Common Finite-Trial Evaluation}
\label{sec:lf-evaluation-audit}

Existing methods evaluate textual memory, search, and policy evolution with
different experience lifetimes and task-feedback opportunities.  \VRLBench
aligns the finite trial budget while preserving each method's defining update.
We examine \Reflexion, two recent textual-update methods, and two influential
search or policy-evolution methods whose papers report a \Reflexion-labeled comparison
\citep{shinn2023reflexion,suzgun2025dynamiccheatsheet,zhang2026ace,
zhou2024lats,zhang2024agentpro}.  \Reflexion enters directly.  We adapt Dynamic
Cheatsheet and ACE to learn from failed attempts within the current task.
LATS and Agent-Pro remain outside the matrix because their defining mechanisms
use additional branch outcomes, search, verification, or development games.
\Cref{tab:lf-post-reflexion-audit} summarizes these choices.

\begin{table}[!htbp]
    \centering
    \footnotesize
    \setlength{\tabcolsep}{2.2pt}
    \renewcommand{\arraystretch}{1.07}
    \begin{tabular*}{\textwidth}{@{\extracolsep{\fill}}
        >{\raggedright\arraybackslash}p{0.13\textwidth}
        >{\raggedright\arraybackslash}p{0.25\textwidth}
        >{\raggedright\arraybackslash}p{0.26\textwidth}
        >{\raggedright\arraybackslash}p{0.27\textwidth}@{}}
        \toprule
        Method & Original experience / opportunity unit &
        Semantics of original \Reflexion comparison & Use in \VRLBench \\
        \midrule
        \Reflexion\newline \citep{shinn2023reflexion} &
        Failed episode $\rightarrow$ reflection $\rightarrow$ reset trial;
        persistent reflection memory across trials. &
        Reference episode-level update. &
        Included directly; the episode-level update fits the failure-only
        finite-trial evaluation. \\
        \addlinespace
        Dynamic Cheatsheet\newline \citep{suzgun2025dynamiccheatsheet} &
        Persistent sheet across independent queries; successes and failures
        may both enter the stream. &
        No experimental \Reflexion row. &
        Included as DC-Cu$^\ast$: the cumulative curator/extractor update is
        retained, but its cross-query experience scope and lifetime are
        replaced by counted failures from the current task-family or goal. \\
        \addlinespace
        ACE\newline \citep{zhang2026ace} &
        Online playbook across a shuffled stream of different tasks; offline
        playbook updated on train before test. &
        No experimental \Reflexion row. &
        Included as ACE$^\ast$: the Reflector and online ADD-only Curator are
        retained, but the original cross-task or train-to-test experience
        lifetime is replaced by counted failures from the current task-family
        or goal. \\
        \addlinespace
        LATS\newline \citep{zhou2024lats} &
        Within-problem Monte Carlo tree search (MCTS); one search can expose
        multiple environment-tested branches, values, simulations, and restored
        states. &
        A \Reflexion-labeled row shares a nominal search counter, not the same
        task-dependent feedback opportunities. &
        Not included; its defining search obtains additional branch outcomes,
        values, simulations, and restored-state access. \\
        \addlinespace
        Agent-Pro\newline \citep{zhang2024agentpro} &
        Policy evolution over train/development games, with search and
        verification where enabled, followed by frozen test evaluation. &
        The \Reflexion-labeled row is pre-action two-pass self-checking, not a
        failed-episode update followed by reset. &
        Not included; its defining policy evolution, search, verification, and
        development data do not fit the common evaluation. \\
        \bottomrule
    \end{tabular*}
    \caption{\textbf{How existing methods enter the common evaluation.}
    \Reflexion enters directly; DC-Cu$^\ast$ and ACE$^\ast$ adapt their textual
    updates to failures within the current task.  LATS and Agent-Pro retain
    search or development mechanisms that require different task-feedback
    opportunities and therefore remain outside the matrix.  Original cohorts,
    hyperparameters, and reported budgets appear in
    \cref{tab:lf-app-source-protocol-details}.}
    \label{tab:lf-post-reflexion-audit}
\end{table}

The asterisk marks the change in experience lifetime for DC-Cu$^\ast$ and
ACE$^\ast$; results apply to these within-task adaptations.  The original
feedback and evaluation conditions, including the search and development
procedures of LATS and Agent-Pro, appear in
\cref{tab:lf-app-source-protocol-details}.

\FloatBarrier

%% file: appendices/evaluation_results.tex
\section{VRL-Bench Evaluation Details and Complete Results}
\label{app:lf-vrlbench}

\subsection{Original Evaluation Settings and VRL-Bench Adaptations}
\label{app:lf-source-protocol-details}

\begin{table}[!htbp]
    \centering
    \footnotesize
    \setlength{\tabcolsep}{1.8pt}
    \renewcommand{\arraystretch}{1.08}
    \begin{tabular*}{\textwidth}{@{\extracolsep{\fill}}
        >{\raggedright\arraybackslash}p{0.12\textwidth}
        >{\raggedright\arraybackslash}p{0.205\textwidth}
        >{\raggedright\arraybackslash}p{0.30\textwidth}
        >{\raggedright\arraybackslash}p{0.125\textwidth}
        >{\raggedright\arraybackslash}p{0.19\textwidth}@{}}
        \toprule
        Method & Original experience and lifetime &
        Original opportunity and evaluation unit &
        Reflexion row / episode-level update? & \VRLBench implementation \\
        \midrule
        \Reflexion\newline \citep{shinn2023reflexion} &
        A failed episode produces a reflection retained across resets;
        original task implementations vary in how much retained memory is
        exposed in the prompt. &
        ALFWorld: 134 tasks, 12 consecutive trials, at most 30 actions each.
        HotPotQA: 100 questions, failed tasks retried until three consecutive
        failures. WebShop: 100 requests; the experiment was stopped after four
        trials when the aggregate curve showed no improvement. &
        Reference / yes &
        \Reflexion: episode-level update included in the common evaluation setting. \\
        \addlinespace
        Dynamic Cheatsheet\newline \citep{suzgun2025dynamiccheatsheet} &
        A persistent sheet is updated across independent queries; successful
        and failed examples can both enter the stream. &
        Claude 3.5 and GPT-4o on AIME, Game of 24, GPQA, Math Equation
        Balancer, and MMLU-Pro.  The unit is the query sequence; reporting is
        not uniformly based on repeated runs with error bars. &
        No / -- &
        DC-Cu$^\ast$: cumulative replace-on-update, restricted to counted
        failures from the current task-family or goal. \\
        \addlinespace
        ACE\newline \citep{zhang2026ace} &
        Online AppWorld follows predict--update over a shuffled test stream of
        different tasks; offline ACE updates on train before pass@1 test. &
        Each online sample is a ReAct episode of at most 40 steps followed by
        Reflector and Curator updates.  Original baselines include base/ReAct,
        ICL, MIPROv2, GEPA, and Dynamic Cheatsheet, but not \Reflexion. &
        No / -- &
        ACE$^\ast$: online ADD-only update, restricted to counted failures
        from the current task-family or goal. \\
        \addlinespace
        LATS\newline \citep{zhou2024lats} &
        An MCTS tree retains action--observation nodes, value estimates,
        terminal simulations, and reflections within the current problem. &
        HotPotQA: 100 questions, GPT-3.5, $k=50$, $n=5$. WebShop: 50
        instructions, depth 15, $k=30$, $n=5$ for LATS; the same nominal $k$
        is used for best-of-$k$ and \Reflexion, but not the same node-level observations.
        Original WebShop results are point estimates without intervals. &
        Yes / not established in the original report &
        Not executed; requires a search, value, simulation, and restoration
        budget. \\
        \addlinespace
        Agent-Pro\newline \citep{zhang2024agentpro} &
        Policies are learned from training games, searched and verified where
        enabled, then frozen for separate test games. &
        Blackjack: 50 failed training games then 900 test games, without DFS or
        verification. Poker: 167 train and 20 development games, eight
        candidate policies per DFS expansion plus verification, then 1,600 test
        games; five repeats are averaged without pairwise intervals. &
        Yes / no &
        Not executed; requires separate policy-search, verification, and data
        lifecycle budgets. \\
        \bottomrule
    \end{tabular*}
    \caption{Original evaluation settings and their VRL-Bench adaptations.
    Experience lifetimes and opportunity units determine what each reported
    score measures.
    ``Reflexion row'' indicates whether the original report includes a row
    under that name, while
    ``episode-level update'' requires environment failure
    before reflection and reset.  $^\ast$ marks a benchmark-constrained
    adaptation; separate-track methods were not executed.}
    \label{tab:lf-app-source-protocol-details}
\end{table}
\Cref{tab:lf-post-reflexion-audit} summarizes which methods we include in the
main comparison.  \Cref{tab:lf-app-source-protocol-details} relates nominal iterations to the
task-dependent observations available in each original setting.

The original \Reflexion settings share the episode--failure--reflection--reset
order, but use task-specific trial limits and context-management choices.  The WebShop experiment was
stopped after trial four when its aggregate curve showed no improvement, rather
than by a formal per-instance stopping rule, and reports no significant
improvement over ReAct.  The original evaluations begin with empty reflection
memory but do not establish exact random-draw or reset-seed reuse across every
method.  These differences motivate the shared-$T_1$ requirement
\citep{shinn2023reflexion}.

Dynamic Cheatsheet evaluates a sheet across independent-query streams.  ACE's
online AppWorld setting updates after each item in a shared shuffled test
stream, while its offline setting updates on training data before pass@1 test.
Neither original evaluation compares these mechanisms in \Reflexion-style
trial and error under a finite complete-trial budget
\citep{suzgun2025dynamiccheatsheet,zhang2026ace}.  For a fair comparison with
\Reflexion and memory-free retry, we retain the cumulative sheet and playbook
updates but restrict experience to counted failures within the current
task-family or goal, reset memory between cases, and match trial and interaction
budgets.  We mark these adaptations with a star.

For LATS, a five-child expansion executes candidates to obtain environment
observations before value evaluation, terminal simulation, reflection, and
executable-state reversion.  Before early stopping, the WebShop parameters in
Algorithm~1 permit up to $30\times15\times5=2{,}250$ candidate A--O edges, while
the original paper reports no realized transition ledger.  Matching the nominal
$k=30$ used for LATS, best-of-$k$, and the row named \Reflexion therefore does
not match task-dependent observations or model and environment calls
\citep{zhou2024lats}.

Agent-Pro learns policies on train/development games before frozen test
evaluation.  In Poker, a failed game may be replayed up to three times during
verification; each DFS expansion generates eight candidates, and each candidate
is evaluated on two development hands under 16 permutations.  A uniform total
game or DFS-node cap is not reported.  The prompt associated with its row named
\Reflexion asks the
same model to check and refine a thought and action before execution, with no
environment outcome, failed trajectory, reset, or cross-episode reflection
between the two responses.  It is therefore a meaningful baseline in the original setting but
not the episode-level procedure used as the present reference
\citep{zhang2024agentpro}.

\FloatBarrier
\subsection{Evaluated Primary Matrix}
\label{app:lf-vrlbench-primary}

The primary matrix crosses three models with two environments:
DeepSeek-V4-Flash, GPT-5.4 nano, and GLM-4.7-FlashX on MiniWoB
and WebShop.  MiniWoB contains 64 task families with 10 paired instances each
($N=640$); WebShop contains the first 100 goals of the official 500-goal human
test split.  Every method receives at most six complete trials and the same
per-episode Action--Observation limit.  Trial 1 is identical across conditions
within each paired model--environment cohort, including the environment
interaction and Actor-visible history; success stops later trials.

\Cref{tab:lf-primary-results} reports the complete five-condition outcome
matrix.  No memory, \Reflexion, \DCCu$^\ast$, and ACE$^\ast$ form the
existing-update comparison; \VEX{} is evaluated separately as the proposed
language-model scheduler.  Across the 2,220 paired cases, the five conditions
yield 11,100 condition--case runs and at most 66,600 condition--trial slots.
Early stopping upon success leaves 16,946 distinct complete-trial executions, with each
shared $T_1$ counted once.

\begin{table}[H]
    \centering
    \footnotesize
    \setlength{\tabcolsep}{3.0pt}
    \renewcommand{\arraystretch}{1.03}
    \textbf{(a) SR@6 (\%), higher is better}\par\smallskip
    \begin{tabular*}{\textwidth}{@{\extracolsep{\fill}}lcccccc@{}}
        \toprule
        \multirow{2}{*}{Method} &
        \multicolumn{3}{c}{MiniWoB ($N=640$)} &
        \multicolumn{3}{c}{WebShop ($N=100$)} \\
        \cmidrule(lr){2-4}\cmidrule(l){5-7}
        & \makecell{DeepSeek-\\V4-Flash}
        & \makecell{GPT-5.4\\nano}
        & \makecell{GLM-4.7-\\FlashX}
        & \makecell{DeepSeek-\\V4-Flash}
        & \makecell{GPT-5.4\\nano}
        & \makecell{GLM-4.7-\\FlashX} \\
        \midrule
        \multicolumn{7}{@{}l}{\itshape Existing verbal updates} \\
        No memory & 87.3 & 82.5 & 82.5 & 45.0 & 39.0 & 50.0 \\
        \Reflexion & 88.9 & 85.9 & 81.2 & 51.0 & 55.0 & 46.0 \\
        \addlinespace[1pt]
        \DCCu$^\ast$ & 85.8 & 81.6 & 79.2 & 46.0 & 45.0 & 39.0 \\
        ACE$^\ast$ & 89.4 & 84.5 & 82.7 & 59.0 & 50.0 & 45.0 \\
        \addlinespace[1pt]
        \multicolumn{7}{@{}l}{\itshape Proposed language-model scheduler} \\
        \VEX{} & 93.9 & 87.0 & 84.1 & 61.0 & 56.0 & 58.0 \\
        \bottomrule
    \end{tabular*}

    \vspace{0.7em}
    \textbf{(b) RR@6 (\%), higher is better}\par\smallskip
    \begin{tabular*}{\textwidth}{@{\extracolsep{\fill}}lcccccc@{}}
        \toprule
        \multirow{2}{*}{Method} &
        \multicolumn{3}{c}{MiniWoB ($N=640$)} &
        \multicolumn{3}{c}{WebShop ($N=100$)} \\
        \cmidrule(lr){2-4}\cmidrule(l){5-7}
        & \makecell{DeepSeek-\\V4-Flash}
        & \makecell{GPT-5.4\\nano}
        & \makecell{GLM-4.7-\\FlashX}
        & \makecell{DeepSeek-\\V4-Flash}
        & \makecell{GPT-5.4\\nano}
        & \makecell{GLM-4.7-\\FlashX} \\
        \midrule
        \multicolumn{7}{@{}l}{\itshape Existing verbal updates} \\
        No memory & 48.4 & 54.3 & 53.5 & 9.8 & 15.3 & 25.4 \\
        \Reflexion & 54.8 & 63.3 & 50.2 & 19.7 & 37.5 & 19.4 \\
        \addlinespace[1pt]
        \DCCu$^\ast$ & 42.0 & 51.8 & 44.8 & 11.5 & 23.6 & 9.0 \\
        ACE$^\ast$ & 56.7 & 59.6 & 53.9 & 32.8 & 30.6 & 17.9 \\
        \addlinespace[1pt]
        \multicolumn{7}{@{}l}{\itshape Proposed language-model scheduler} \\
        \VEX{} & 75.2 & 66.1 & 57.7 & 36.1 & 38.9 & 37.3 \\
        \bottomrule
    \end{tabular*}

    \vspace{0.7em}
    \textbf{(c) Censored AvgT@6, lower is better}\par\smallskip
    \begin{tabular*}{\textwidth}{@{\extracolsep{\fill}}lcccccc@{}}
        \toprule
        \multirow{2}{*}{Method} &
        \multicolumn{3}{c}{MiniWoB ($N=640$)} &
        \multicolumn{3}{c}{WebShop ($N=100$)} \\
        \cmidrule(lr){2-4}\cmidrule(l){5-7}
        & \makecell{DeepSeek-\\V4-Flash}
        & \makecell{GPT-5.4\\nano}
        & \makecell{GLM-4.7-\\FlashX}
        & \makecell{DeepSeek-\\V4-Flash}
        & \makecell{GPT-5.4\\nano}
        & \makecell{GLM-4.7-\\FlashX} \\
        \midrule
        \multicolumn{7}{@{}l}{\itshape Existing verbal updates} \\
        No memory & 1.99 & 2.46 & 2.40 & 4.41 & 4.81 & 4.39 \\
        \Reflexion & 1.92 & 2.25 & 2.46 & 4.14 & 4.19 & 4.46 \\
        \addlinespace[1pt]
        \DCCu$^\ast$ & 2.05 & 2.48 & 2.58 & 4.40 & 4.74 & 4.84 \\
        ACE$^\ast$ & 1.91 & 2.33 & 2.41 & 3.91 & 4.40 & 4.53 \\
        \addlinespace[1pt]
        \multicolumn{7}{@{}l}{\itshape Proposed language-model scheduler} \\
        \VEX{} & 1.75 & 2.23 & 2.37 & 3.76 & 4.17 & 4.08 \\
        \bottomrule
    \end{tabular*}
    \caption{Complete \VRLBench outcomes across five conditions and six
    settings.  The panels report SR@6, RR@6, and AvgT@6 under the common trial
    budget and per-trial A--O limits.  $^\ast$ marks failure-only adaptations
    of original update mechanisms.  Paired effects are reported in
    \cref{tab:vrlbench-reevaluation-paired,tab:lf-app-vrlbench-inference}.}
\label{tab:lf-primary-results}
\end{table}

\subsection{Existing-Update Comparison: Paired Inference}
\label{app:lf-vrlbench-reevaluation-inference}

\Cref{tab:vrlbench-reevaluation-paired} reports the within-task comparison in
\cref{sec:lf-reevaluation}.  Relative to memory-free retry, \Reflexion has
four positive and two negative point
estimates; one positive mean-effect interval excludes zero, GPT-5.4 nano on
WebShop at $+16$ points $[+8,+25]$.  DC-Cu$^\ast$ has two positive and four
negative point estimates.  Neither positive interval excludes zero, while two
negative intervals do: GLM-4.7-FlashX on MiniWoB at $-3.3$ points
$[-6.3,-0.3]$ and on WebShop at $-11$ points $[-19,-3]$.  ACE$^\ast$
has five positive and one negative point estimate, with positive mean-effect
intervals excluding zero for DeepSeek-V4-Flash on MiniWoB at $+2.0$ points
$[+0.2,+4.4]$, DeepSeek-V4-Flash on WebShop at $+14$ points $[+7,+22]$, and
GPT-5.4 nano on WebShop at $+11$ points $[+2,+20]$.  For the DeepSeek-V4-Flash
MiniWoB ACE contrast, the mean-effect interval excludes zero while the exact
sign test gives $p=.227$.  The sign test counts the direction of non-tied
task-family differences rather than their mean magnitude.

All 18 effects are paired contrasts.  MiniWoB uses
64 paired task-family means and exact sign tests, WebShop uses 100 paired goals
and exact McNemar tests, and the intervals use 100,000 paired bootstrap
resamples with seed 0.  Each statement is setting-specific: effects are not
pooled, no multiplicity adjustment is applied, and an interval that includes zero
leaves direction unresolved rather than establishing equality.
Throughout the paper, intervals quantify uncertainty across evaluation units
in the recorded runs; they do not capture variation from fresh model calls or
regenerated environment instances.

\input{tables/paired_inference.tex}

\FloatBarrier
\subsection{Environment Feedback and Method-Internal Records}
\label{app:lf-vrlbench-information-flow}

The version-pinned harness records two objects after an eligible counted failure.  The
benchmark-visible policy record contains the initial observation, every dispatched
action and its visible observation, and the public terminal close record.  Its
eligibility bit is $\eta_t^{(m)}=1$ exactly when that valid record
contains at least one dispatched A--O pair.  Eligible failed records form the
ordered failure history $\mathcal F_t^{(m)}$.  Separately, the harness may
retain a method-internal Actor record $q_t^{(m)}$: raw model responses,
Thought/Action text, parse or format feedback, and candidates that were
generated but not dispatched.  The latter is endogenous computation from the
same eligible counted failure, not another interaction opportunity.  Any
external observation reproduced in it must be a reference to or projection of
the initial observation, an executed A--O pair, or the public outcome already
present in the benchmark-visible record; it cannot introduce an extra rollout,
verifier, evaluator, oracle, or uncounted interaction.

Eligibility distinguishes an executed interaction with Actor-visible feedback
from text that never reached the environment.  An action rejected before
dispatch is not an environment transition, and internal reasoning, an
unexecuted candidate, or a parser rejection cannot create an environment
observation.  By contrast, an invalid
action that reaches the environment and produces Actor-visible feedback is an
executed A--O pair.  The same counted trial may therefore contain endogenous
text that is retained in $q_t^{(m)}$ without promoting that text to an
environment fact.

A zero-dispatch protocol failure has $\eta_t^{(m)}=0$.  It remains a counted
complete trial in the run and resource ledgers, reduces the remaining trial budget,
and enters neither $\mathcal F_t^{(m)}$ nor $\mathcal Q_t^{(m)}$.  Its public
initial observation and close reason remain in the trial ledger for resource
accounting and reconstruction, but
they do not become cross-trial environment feedback.  The evaluation runner
therefore skips the cross-trial writer and failure hook for that close.  When
another trial remains, it still follows the method's defined read/build
lifecycle: persistent memory may
remain unchanged, while a method that conditions on $R_t$ may recompute context
from the unchanged failure history and its smaller value.  For \VEX{}, this means
calling the scheduler with the unchanged forest and smaller $R_t$ rather than
treating the protocol failure as environment feedback.

The exactly shared memory-free $T_1$ includes $e_1$, its available method-internal
$q_1$, and $\eta_1$.  If the opening trial is an eligible failure, those
records become available to each method's declared projections, although
$Q_m$ may be empty.  If it is ineligible, neither record enters
$\mathcal F_1$ or $\mathcal Q_1$ for any method; the shared trial still
consumes the first opportunity.  Later external and endogenous records may
diverge because the updates induce different policies.  What remains matched
is the opening realization, eligibility rule, external-state schema, and
complete-trial budget, not an artificial identity of later histories.

The distinction matters for method fidelity.  MiniWoB \Reflexion uses its
method-internal Actor decision-block record, whereas WebShop \Reflexion
reconstructs its writer input from the benchmark-visible executed record.  \DCCu$^\ast$
and ACE$^\ast$ use the method-internal Actor record in both
environments.  \VEX{} uses only the
benchmark-visible record; No memory performs no update.

The runner populates environment-feedback messages from the same
initial observation or dispatched A--O step written to the benchmark-visible record.
Parser feedback is marked before any environment dispatch, and WebShop's
private target and reward-detail fields are removed before public terminal
feedback is stored.  The version-pinned implementation serializes the method-internal Actor
record as labelled flat text, preserving the original writer view.  Because an
Actor can itself emit text resembling an observation label, we use the benchmark-visible
record as the only source of external observations and treat the
method-internal record as model-generated text.

\subsection{\VEX{} execution loop}
\label{app:lf-vrlbench-dsst-loop}

\begin{algorithm}[H]
\caption{\VEX{} inter-episode loop inside \VRLBench}
\label{alg:lf-dsst-loop}
\begin{algorithmic}[1]
\Require task $x$; public rules $\rho$; scheduler model
$\mathcal{S}_{\theta}$; shared Actor configuration $A$; no parameter updates;
environment
$\mathrm{Env}$; complete-trial budget $T$
\Statex \textbf{State:} $\mathcal{F}_0\gets[\,]$;
$\mathcal{D}_0\gets\varnothing$
\For{$t=1,\ldots,T$}
    \State reset $\mathrm{Env}$
    \State $(Y_t,e_t,\eta_t)\gets
    A(x,\rho,\mathcal{D}_{t-1};\mathrm{Env})$
    \Comment{$\mathcal{D}_{t-1}$ enters verbatim}
    \If{$Y_t=1$}
        \State \Return \textsc{Success}
    \EndIf
    \State $\mathcal{F}_t\gets\mathcal{F}_{t-1}$
    \If{$\eta_t=1$}
        \State append $(t,e_t)$ to $\mathcal{F}_t$
    \EndIf
    \If{$t<T$}
        \State $R_t\gets T-t$
        \State $\mathcal{D}_t\gets
        \mathcal{S}_{\theta}(x,\rho,\mathcal{F}_t,R_t)$
        \Comment{one call; forest unchanged if $\eta_t=0$}
    \EndIf
\EndFor
\State \Return \textsc{Failure}
\end{algorithmic}
\end{algorithm}

The outer loop in \cref{alg:lf-dsst-loop} makes the eligibility and budget
semantics explicit.
The first trial has no cross-trial decision object.  A failed eligible record
extends the forest; a zero-dispatch protocol failure leaves it unchanged.
Both are counted trials, so both reduce the remaining trial budget.  Whenever a
further trial remains, \VEX{} forms one new raw response from the current forest
and $R_t$ and passes it unchanged to the next Actor.

The scheduler call after an ineligible failure is not a failure-history
update.  It is \VEX{}'s context construction under a
smaller public remaining budget.  Malformed, contradictory, or truncated model output
remains counted model behavior; a transport failure with no returned response
is recorded separately and is never treated as environment feedback.

\subsection{Worked trajectory pair}
\label{app:lf-vrlbench-worked-record}

\begin{figure}[!htbp]
    \centering
    \includegraphics[width=0.98\textwidth]{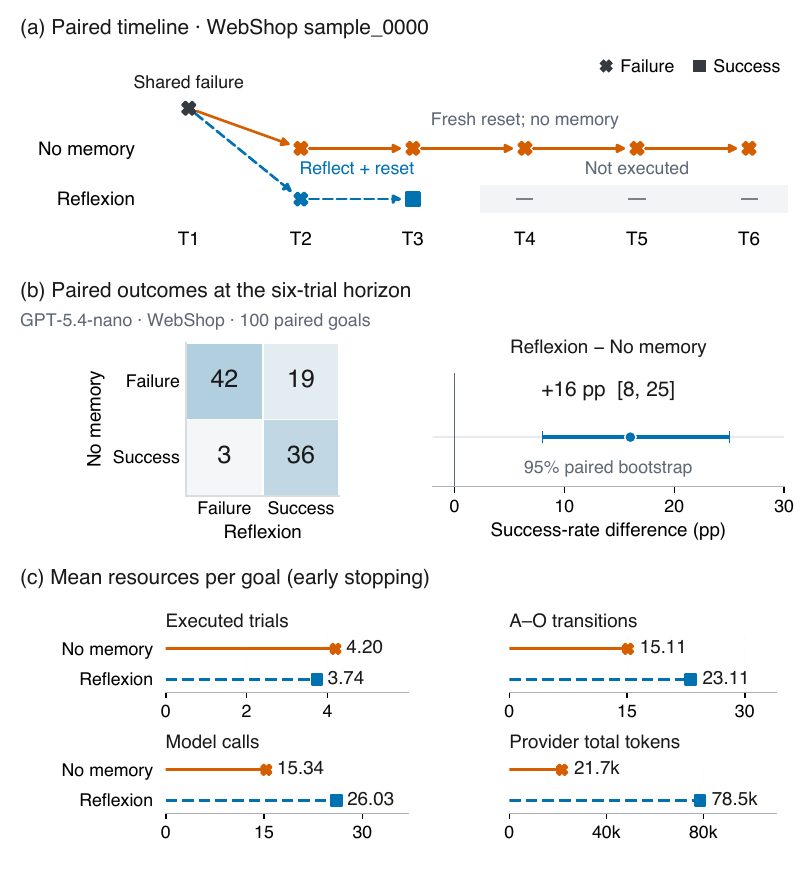}
    \caption{\textbf{A worked pair links one process example to its
    setting-level effect and realized resource use.}  (a) In the
    outcome-independently selected
    GPT-5.4-nano WebShop \texttt{sample\_0000}, both conditions share the same
    opening failure; No memory then fails through $T_6$, whereas \Reflexion
    succeeds at $T_3$ and stops.  (b) Across 100 paired goals, \Reflexion minus
    No memory is $+16$ SR@6 points with a 95\% paired-bootstrap interval of
    $[+8,+25]$.  (c) Trials, A--O transitions, calls, and tokens report the
    realized resource use with early stopping upon success.}
    \label{fig:lf-vrlbench-paired-experiment}
\end{figure}

In \cref{fig:lf-vrlbench-paired-experiment}, the event history explains the
update-and-reset sequence, while the setting-level contingency table estimates
its aggregate outcome difference.  The resource axes show how early stopping upon success
changes realized trials, transitions, calls, and tokens in their respective units.

\FloatBarrier
\subsection{\VEX{} paired inference}
\label{app:lf-vrlbench-inference}

We compare the proposed \VEX{} scheduler with both memory-free retry and
\Reflexion.  These paired comparisons evaluate the same integrated \VEX{}
condition separately from the existing-update re-evaluation in
\cref{tab:vrlbench-reevaluation-paired}.

\begin{table}[!htb]
\centering
\scriptsize
\setlength{\tabcolsep}{3.6pt}
\renewcommand{\arraystretch}{1.06}
\begin{tabular*}{\textwidth}{@{\extracolsep{\fill}}llrrrlrr@{}}
\toprule
Benchmark & Comparator $\rightarrow$ \VEX{} & Base & \VEX{} & $\Delta$ pp &
95\% CI & $+/-$ & $p$ \\
\midrule
\multicolumn{8}{@{}l}{\textit{DeepSeek-V4-Flash}} \\
MiniWoB & No memory & 87.3 & 93.9 & +6.6 & [+2.19,+12.03] & 11/1 & 0.0063 \\
MiniWoB & \Reflexion & 88.9 & 93.9 & +5.0 & [+1.88,+8.75] & 11/1 & 0.0063 \\
WebShop & No memory & 45.0 & 61.0 & +16.0 & [+9.00,+23.00] & 16/0 & $3.05\!\times\!10^{-5}$ \\
WebShop & \Reflexion & 51.0 & 61.0 & +10.0 & [+5.00,+16.00] & 10/0 & 0.0020 \\
\addlinespace[2pt]
\multicolumn{8}{@{}l}{\textit{GPT-5.4 nano}} \\
MiniWoB & No memory & 82.5 & 87.0 & +4.5 & [+0.47,+9.38] & 9/4 & 0.267 \\
MiniWoB & \Reflexion & 85.9 & 87.0 & +1.1 & [-0.94,+3.75] & 7/5 & 0.774 \\
WebShop & No memory & 39.0 & 56.0 & +17.0 & [+10.00,+25.00] & 17/0 & $1.53\!\times\!10^{-5}$ \\
WebShop & \Reflexion & 55.0 & 56.0 & +1.0 & [-8.00,+10.00] & 11/10 & 1.000 \\
\addlinespace[2pt]
\multicolumn{8}{@{}l}{\textit{GLM-4.7-FlashX}} \\
MiniWoB & No memory & 82.5 & 84.1 & +1.6 & [-1.09,+4.38] & 10/9 & 1.000 \\
MiniWoB & \Reflexion & 81.2 & 84.1 & +2.8 & [+0.78,+5.16] & 13/4 & 0.049 \\
WebShop & No memory & 50.0 & 58.0 & +8.0 & [-1.00,+17.00] & 15/7 & 0.134 \\
WebShop & \Reflexion & 46.0 & 58.0 & +12.0 & [+2.00,+22.00] & 19/7 & 0.029 \\
\bottomrule
\end{tabular*}
\caption{Paired SR@6 inference for the proposed \VEX{} scheduler.  MiniWoB
uses 64 paired task-family differences, bootstrap intervals, and exact sign
tests; WebShop uses 100 paired goals, bootstrap intervals, and exact McNemar
tests.  Intervals use 100,000 resamples (seed 0).  The $+/-$ column counts paired
task families (MiniWoB) or goals (WebShop) favoring \VEX{} or the comparator,
respectively; ties are omitted.}
\label{tab:lf-app-vrlbench-inference}
\end{table}

Against memory-free retry, all six point estimates favor \VEX{} and four of the
six setting-specific paired intervals exclude zero; both GLM-4.7-FlashX
intervals include zero.  Against \Reflexion, all six point estimates again
favor \VEX{} and four intervals exclude zero; both GPT-5.4-nano intervals include
zero.  These are setting-specific, unadjusted paired comparisons.

\FloatBarrier
\subsection{Recovery after the first-trial failure}
\label{app:lf-vrlbench-recovery}

\begin{table}[!htb]
\centering
\footnotesize
\setlength{\tabcolsep}{2.5pt}
\begin{tabular*}{\textwidth}{@{\extracolsep{\fill}}lrrrr@{}}
\toprule
Setting & \makecell{Shared $T_1$\\failures} &
\makecell{No memory\\recovered} & \makecell{\Reflexion\\recovered} &
\makecell{\VEX{}\\recovered} \\
\midrule
DeepSeek-V4-Flash / MiniWoB & 157 & 76 & 86 & 118 \\
DeepSeek-V4-Flash / WebShop & 61 & 6 & 12 & 22 \\
GPT-5.4 nano / MiniWoB & 245 & 133 & 155 & 162 \\
GPT-5.4 nano / WebShop & 72 & 11 & 27 & 28 \\
GLM-4.7-FlashX / MiniWoB & 241 & 129 & 121 & 139 \\
GLM-4.7-FlashX / WebShop & 67 & 17 & 13 & 25 \\
\bottomrule
\end{tabular*}
\caption{Post-$T_1$ recovery counts for Retry, \Reflexion, and \VEX{} in each
setting.  The denominator is the exactly shared first-trial failure cohort;
paired uncertainty is reported in
\Cref{tab:vrlbench-reevaluation-paired,tab:lf-app-vrlbench-inference}.}
\label{tab:lf-app-vrlbench-recovery}
\end{table}

The recovery totals in \cref{fig:lf-vrlbench-recovery} locate the terminal
success differences.  The next analysis separates these recoveries from
changes in time to success.

\FloatBarrier
\subsection{Separating recovery from time to success}
\label{app:lf-vrlbench-trial-timing}

Censored AvgT@6 combines terminal success and its timing.  In particular,
with SR@$t$ expressed as a proportion,
\begin{equation}
  \mathrm{AvgT@6}=7-\sum_{t=1}^{6}\mathrm{SR@}t.
\end{equation}
We decompose the \VEX{}-minus-Retry difference using the first-success trials
$T_{i,D}^{\star}$ and $T_{i,R}^{\star}$, assigning unsolved cases to 7.
Let $G$ contain cases solved by exactly one method and $B$ those solved by both.
Then
\begin{equation}
\Delta\mathrm{AvgT@6}
=\frac{1}{N}\sum_{i\in G}(T_{i,D}^{\star}-T_{i,R}^{\star})
+\frac{1}{N}\sum_{i\in B}(T_{i,D}^{\star}-T_{i,R}^{\star}).
\end{equation}
The first term captures gains and losses in terminal recovery; the second
captures timing among common successes.  Cases unsolved by both contribute
zero.  Both terms use the full cohort denominator, so they add to the reported
overall difference rather than comparing differently sized subgroup means.

\input{tables/trial_timing.tex}

The recovery term dominates in all six settings.  On GLM-4.7-FlashX MiniWoB,
the 513 cases solved by both methods require 18 more trials in total under \VEX{},
contributing $18/640=+0.028$ to AvgT@6; additional recoveries more than offset
this timing difference.  These outcome-defined groups describe the observed
trajectories rather than isolating a causal effect on completion speed.

With early stopping upon success, the mean number of executed trials is
$\mathrm{AvgT@6}-(1-\mathrm{SR@6})$.  Its \VEX{}-minus-Retry differences are
$-0.169$, $-0.181$, and $-0.013$ on MiniWoB, and $-0.490$, $-0.470$, and
$-0.230$ on WebShop, in DeepSeek-V4-Flash, GPT-5.4 nano, and GLM-4.7-FlashX
order.  Lower trial use does not imply fewer model calls or tokens; those
resources are reported separately below.

\subsection{Realized resource use}
\label{app:lf-vrlbench-resources}

The matched outer opportunity unit is the complete trial; the per-trial A--O
limit is also held fixed.  Provider tokens and model calls are not matched.  The
tables below report realized means with early stopping upon success.  Each compact cell
is environment transitions / model calls / provider-reported
prompt-plus-completion tokens per instance.

\begin{table}[H]
\centering
\scriptsize
\setlength{\tabcolsep}{5pt}
\renewcommand{\arraystretch}{1.08}
\begin{tabular*}{\textwidth}{@{\extracolsep{\fill}}lccc@{}}
\toprule
MiniWoB method & DeepSeek-V4-Flash & GPT-5.4 nano & GLM-4.7-FlashX \\
\midrule
No memory & 7.42 / 2.79 / 5,481 & 8.47 / 3.18 / 5,847 & 10.44 / 3.77 / 6,902 \\
\Reflexion & 5.97 / 2.89 / 6,700 & 7.50 / 4.01 / 9,849 & 10.55 / 4.93 / 10,622 \\
\DCCu$^\ast$ & 7.83 / 3.54 / 11,668 & 8.41 / 4.44 / 13,225 & 10.99 / 5.10 / 15,074 \\
ACE$^\ast$ & 6.07 / 3.43 / 8,529 & 6.54 / 5.05 / 13,370 & 9.26 / 5.75 / 13,403 \\
\VEX{} & 6.90 / 2.86 / 14,089 & 8.51 / 4.13 / 23,932 & 10.15 / 4.59 / 26,047 \\
\bottomrule
\end{tabular*}
\caption{Complete realized MiniWoB resource ledger.  Cell order is
transitions / calls / tokens; all values are means per instance.}
\label{tab:lf-app-vrlbench-resources-miniwob}
\end{table}

\begin{table}[H]
\centering
\scriptsize
\setlength{\tabcolsep}{5pt}
\renewcommand{\arraystretch}{1.08}
\begin{tabular*}{\textwidth}{@{\extracolsep{\fill}}lccc@{}}
\toprule
WebShop method & DeepSeek-V4-Flash & GPT-5.4 nano & GLM-4.7-FlashX \\
\midrule
No memory & 18.79 / 18.79 / 39,118 & 15.11 / 15.34 / 21,703 & 21.51 / 21.73 / 46,880 \\
\Reflexion & 30.11 / 32.76 / 120,545 & 23.11 / 26.03 / 78,506 & 29.67 / 32.95 / 96,955 \\
\DCCu$^\ast$ & 21.19 / 24.06 / 98,671 & 16.02 / 19.36 / 50,315 & 22.90 / 27.76 / 109,954 \\
ACE$^\ast$ & 34.06 / 39.13 / 155,556 & 24.98 / 31.20 / 92,106 & 30.02 / 38.24 / 117,709 \\
\VEX{} & 21.52 / 23.93 / 94,558 & 14.66 / 17.50 / 52,291 & 19.38 / 23.03 / 74,602 \\
\bottomrule
\end{tabular*}
\caption{Complete realized WebShop resource ledger.  Cell order is
transitions / calls / tokens; all values are means per instance.}
\label{tab:lf-app-vrlbench-resources-webshop}
\end{table}

\begin{figure}[H]
    \centering
    \includegraphics[width=0.96\textwidth]{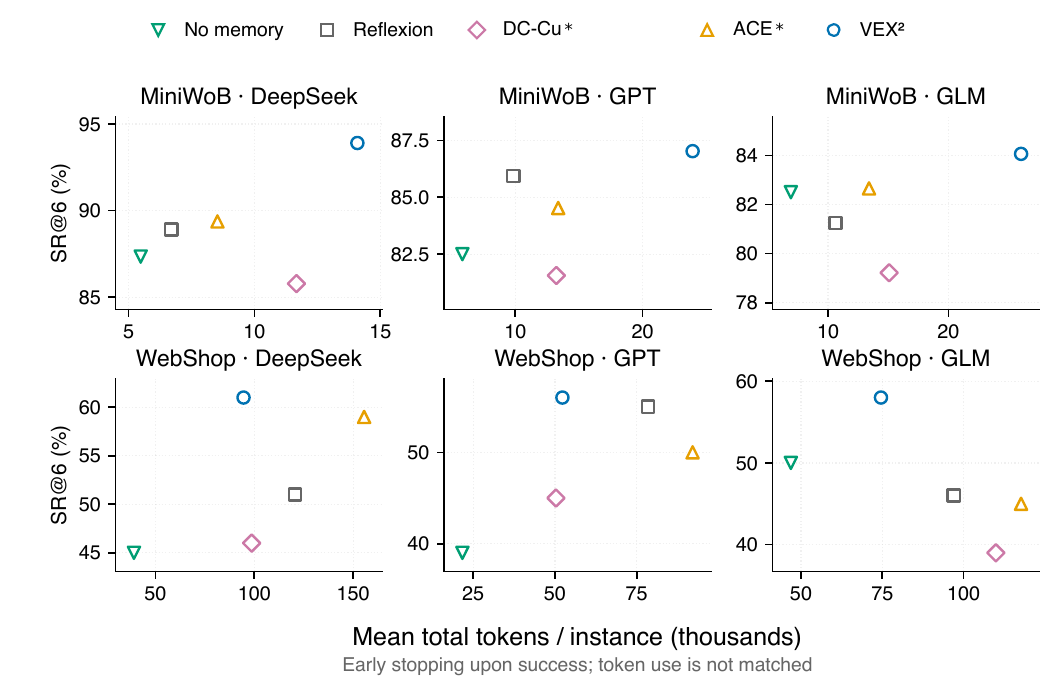}
    \caption{\textbf{Success rates and token use do not follow the same ranking.}
    Across the six evaluated settings, the highest-SR@6 method is
    not uniformly the largest token user, and \DCCu$^\ast$ uses more tokens
    than \Reflexion while attaining lower SR@6 in all three MiniWoB panels.
    Tokens measure realized use with early stopping upon success rather than a matched budget.}
    \label{fig:lf-token-score}
\end{figure}

Early stopping upon success makes realized resource use outcome-dependent.
\Cref{fig:lf-token-score} compares observed success and token use; the latter
is measured rather than matched across conditions.

\subsection{Additional GLM-5.1 MiniWoB Cohort}
\label{app:lf-vrlbench-glm51}

The GLM-5.1 MiniWoB run uses a separate paired cohort: reset-seed episode
indices 10--19 rather than the 20--29 indices used in the primary settings.
We therefore report it separately from the primary $3\times2$ matrix; it does
not isolate the effect of changing the model backbone.

\begin{table}[H]
\centering
\small
\setlength{\tabcolsep}{7pt}
\begin{tabular*}{0.76\textwidth}{@{\extracolsep{\fill}}lrrr@{}}
\toprule
Method & SR@6 & RR@6 & AvgT@6 \\
\midrule
No memory & 89.1 & 41.7 & 1.80 \\
\Reflexion & 90.5 & 49.2 & 1.80 \\
\DCCu$^\ast$ & 88.9 & 40.8 & 1.84 \\
ACE$^\ast$ & 91.7 & 55.8 & 1.71 \\
\VEX{} & 94.2 & 69.2 & 1.63 \\
\bottomrule
\end{tabular*}
\caption{Additional GLM-5.1 results on all 64 MiniWoB task
families with paired episode indices 10--19 ($N=640$, $T=6$).}
\label{tab:lf-app-vrlbench-glm51-outcomes}
\end{table}

\begin{table}[H]
\centering
\small
\setlength{\tabcolsep}{6pt}
\begin{tabular*}{0.78\textwidth}{@{\extracolsep{\fill}}lrrrr@{}}
\toprule
Metric & \Reflexion & \VEX{} & Paired $\Delta$ & 95\% CI \\
\midrule
SR@6 & 90.5 & 94.2 & +3.8 & [+1.3,+6.7] \\
RR@6 & 49.2 & 69.2 & +20.0 & [+8.7,+31.2] \\
AvgT@6 & 1.80 & 1.63 & -0.17 & [-0.30,-0.06] \\
\bottomrule
\end{tabular*}
\caption{Within-setting paired effects for the secondary GLM-5.1 cohort.  SR
and RR differences are
percentage points; negative AvgT is better.  Intervals use 100,000 paired
task-family bootstrap resamples (seed 0).  For SR@6, 8 non-tied families favor
\VEX{} and none favor \Reflexion (exact sign $p=0.0078$).}
\label{tab:lf-app-vrlbench-glm51-inference}
\end{table}

\VEX{} is $+5.1$ SR@6 points above No memory in this additional cohort.  Because
we did not compute a paired \VEX{}-minus-retry interval and the cohort differs
from the primary matrix, this result is not included in the six-setting claim.

%% file: tables/paired_inference.tex
\begin{table}[H]
\centering
\footnotesize
\setlength{\tabcolsep}{3.2pt}
\renewcommand{\arraystretch}{1.04}
\begin{tabular*}{\textwidth}{@{\extracolsep{\fill}}llrrrr@{}}
\toprule
Benchmark & Update $-$ No memory & $\Delta$ (pp) & 95\% CI & $+/-$ (ties) & Exact $p$ \\
\midrule
\multicolumn{6}{@{}l}{\textit{DeepSeek-V4-Flash}} \\
MiniWoB & \Reflexion & +1.6 & [-0.6, +4.1] & 5/3 (56) & 0.7266 \\
MiniWoB & DC-Cu$^{\ast}$ & -1.6 & [-5.5, +1.7] & 5/4 (55) & 1 \\
MiniWoB & ACE$^{\ast}$ & +2.0 & [+0.2, +4.4] & 8/3 (53) & 0.2266 \\
WebShop & \Reflexion & +6.0 & [0.0, +12.0] & 8/2 (90) & 0.1094 \\
WebShop & DC-Cu$^{\ast}$ & +1.0 & [-6.0, +8.0] & 6/5 (89) & 1 \\
WebShop & ACE$^{\ast}$ & +14.0 & [+7.0, +22.0] & 15/1 (84) & $5.19\!\times\!10^{-4}$ \\
\addlinespace[3pt]
\multicolumn{6}{@{}l}{\textit{GPT-5.4 nano}} \\
MiniWoB & \Reflexion & +3.4 & [-0.3, +8.1] & 7/5 (52) & 0.7744 \\
MiniWoB & DC-Cu$^{\ast}$ & -0.9 & [-3.1, +1.7] & 5/13 (46) & 0.0963 \\
MiniWoB & ACE$^{\ast}$ & +2.0 & [-0.9, +5.5] & 6/5 (53) & 1 \\
WebShop & \Reflexion & +16.0 & [+8.0, +25.0] & 19/3 (78) & $8.55\!\times\!10^{-4}$ \\
WebShop & DC-Cu$^{\ast}$ & +6.0 & [-2.0, +14.0] & 11/5 (84) & 0.2101 \\
WebShop & ACE$^{\ast}$ & +11.0 & [+2.0, +20.0] & 16/5 (79) & 0.0266 \\
\addlinespace[3pt]
\multicolumn{6}{@{}l}{\textit{GLM-4.7-FlashX}} \\
MiniWoB & \Reflexion & -1.3 & [-3.3, +0.8] & 3/11 (50) & 0.0574 \\
MiniWoB & DC-Cu$^{\ast}$ & -3.3 & [-6.3, -0.3] & 4/18 (42) & 0.0043 \\
MiniWoB & ACE$^{\ast}$ & +0.2 & [-1.7, +2.2] & 4/5 (55) & 1 \\
WebShop & \Reflexion & -4.0 & [-12.0, +4.0] & 6/10 (84) & 0.4545 \\
WebShop & DC-Cu$^{\ast}$ & -11.0 & [-19.0, -3.0] & 3/14 (83) & 0.0127 \\
WebShop & ACE$^{\ast}$ & -5.0 & [-13.0, +3.0] & 6/11 (83) & 0.3323 \\
\bottomrule
\end{tabular*}
\caption{Controlled re-evaluation relative to memory-free retry. DC-Cu$^{\ast}$ denotes the cumulative-replacement failure-only adaptation; ACE$^{\ast}$ denotes the ADD-only failure-only adaptation. MiniWoB intervals resample 64 paired task-family mean differences (10 instances per family); WebShop intervals resample 100 paired goals. All intervals use 100{,}000 paired percentile-bootstrap resamples with seed 0. The $+/-$ columns count positive and negative paired units, with ties in parentheses; $p$ is an exact sign test for MiniWoB and an exact McNemar test for WebShop. A mean-effect interval and a direction-only exact test need not give the same evidential summary. Positive values favor the named verbal update and negative values favor No memory. Estimates are setting-specific; no pooled effect or multiplicity-adjusted interval is reported.}
\label{tab:vrlbench-reevaluation-paired}
\end{table}

%% file: tables/trial_timing.tex
\begin{table}[!htbp]
\centering
\small
\setlength{\tabcolsep}{3pt}
\begin{tabular*}{\textwidth}{@{\extracolsep{\fill}}llrrr@{}}
\toprule
Benchmark & Model & \shortstack{Recovery\\contribution} & \shortstack{Both-success\\timing contribution} & $\Delta$AvgT@6 \\
\midrule
MiniWoB & DeepSeek-V4-Flash & -0.220 & -0.014 & -0.234 \\
MiniWoB & GPT-5.4 nano & -0.181 & -0.045 & -0.227 \\
MiniWoB & GLM-4.7-FlashX & -0.056 & +0.028 & -0.028 \\
WebShop & DeepSeek-V4-Flash & -0.640 & -0.010 & -0.650 \\
WebShop & GPT-5.4 nano & -0.630 & -0.010 & -0.640 \\
WebShop & GLM-4.7-FlashX & -0.310 & +0.000 & -0.310 \\
\bottomrule
\end{tabular*}
\caption{Descriptive post-hoc decomposition of \VEX{} minus Retry censored AvgT@6. Recovery combines \VEX{}-only and Retry-only successes; both-success timing compares cases solved by both methods. Contributions use the full cohort denominator (640 MiniWoB cases or 100 WebShop goals); both-failure cases contribute zero. Negative values favor \VEX{}. Entries are rounded independently. Outcome-defined groups do not identify causal effects.}
\label{tab:lf-dsst-trial-timing}
\end{table}

%% file: appendices/replay.tex
\section{Replay-Exposure Interventions and Case-Level Analysis}
\label{app:lf-replay}

\subsection{Replay-Exposure Design}
\label{app:lf-replay-design}

We control how often an already-written reflection returns to the same
fixed-parameter Actor in \Reflexion's canonical HotPotQA and AlfWorld settings
\citep{shinn2023reflexion}.  The gate changes only whether prior reflections
appear in the next Actor context, allowing us to measure sensitivity to replay
schedule under a finite trial budget.

Let $\mathcal{M}_t$ be the ordered reflection memory
after complete trial $t$, and let $\mathcal{C}^{0}_{t+1}$ be the next Actor
context without that memory.  For each interior schedule, following every
eligible failure the self-reflection model updates $\mathcal{M}_t$; the next context is then
determined by
\[
G_{t+1}\sim\operatorname{Bernoulli}(1-\varepsilon),\qquad
\mathcal{C}_{t+1}=
\begin{cases}
\mathcal{C}^{0}_{t+1}\oplus\mathcal{M}_t,&G_{t+1}=1,\\
\mathcal{C}^{0}_{t+1},&G_{t+1}=0.
\end{cases}
\]
Thus $\varepsilon=0$ is full replay and $\varepsilon=1$ is the no-replay
condition, with no reflection visible to the next Actor.  The reported grid reuses
the full-replay and no-memory runs for these two endpoints; the no-memory
endpoint supplies no reflection to the Actor.  Interior schedules retain
reflection writing and randomize its exposure.  Trial 1 is shared and
memory-free; terminal success stops
further trials.  The five fixed gates
$\{0,.25,.50,.75,1\}$ yield 6,520 condition--case--trial opportunities.
Early stopping upon success leaves 3,324 recorded
condition--trial episodes, counting the shared first-trial record in each
condition rather than as distinct executions.

Appendix~\ref{app:iclr-replay-overview} defines the conditional recovery rate
and fixed-budget AUC accompanying the cumulative success curves.

\subsection{Full exposure grid and trial-index curves}
\label{app:lf-replay-grid}

\begin{table}[H]
\centering
\small
\setlength{\tabcolsep}{4.8pt}
\begin{tabular*}{\textwidth}{@{\extracolsep{\fill}}lrrrrr@{}}
\toprule
& \multicolumn{5}{c}{Withholding probability $\varepsilon$} \\
\cmidrule(lr){2-6}
Benchmark & 0 & .25 & .50 & .75 & 1 \\
\midrule
HotPotQA & 49.0 & 45.0 & 49.0 & 46.0 & \textbf{50.0} \\
AlfWorld & 76.1 & \textbf{81.3} & 77.6 & 76.1 & 74.6 \\
\bottomrule
\end{tabular*}
\caption{Terminal success (\%) for the complete fixed-$\varepsilon$
grid with GPT-4o mini.  These are observed cohort point estimates.  Interior
schedules retain reflection writing and vary Actor-side exposure; the two
endpoints reuse the full-replay and no-memory runs.}
\label{tab:lf-app-replay-grid}
\end{table}

\begin{table}[H]
\centering
\small
\setlength{\tabcolsep}{5.0pt}
\begin{tabular*}{\textwidth}{@{\extracolsep{\fill}}lrrrrrr@{}}
\toprule
Condition & $T_1$ & $T_2$ & $T_3$ & $T_4$ & $T_5$ & $T_6$ \\
\midrule
\multicolumn{7}{@{}l}{\textit{HotPotQA}} \\
Full withholding & 36.0 & 42.0 & 44.0 & 47.0 & 50.0 & -- \\
Full replay & 36.0 & 43.0 & 46.0 & 48.0 & 49.0 & -- \\
\addlinespace[2pt]
\multicolumn{7}{@{}l}{\textit{AlfWorld}} \\
Full withholding & 56.7 & 64.2 & 67.2 & 70.9 & 71.6 & 74.6 \\
Full replay & 56.7 & 68.7 & 76.1 & 76.1 & 76.1 & 76.1 \\
\bottomrule
\end{tabular*}
\caption{Cumulative success by complete trial (\%) for the full-replay and
full-withholding endpoints.}
\label{tab:lf-app-replay-curves}
\end{table}

The complete cumulative-success and conditional-recovery profiles appear in
\cref{fig:lf-replay-dynamics}.  Because success stops an instance, conditional
recovery is measured among the cases that remain unsolved at each trial.

The paired comparisons accompanying the observed grid are reported in
\Cref{tab:lf-replay-inference}.  The two AlfWorld rows use simultaneous
paired-case max-absolute-bootstrap intervals over every displayed interior
schedule against both endpoints; the HotPotQA endpoint contrast was not an
interior schedule selected from the grid and uses a pointwise paired-case
percentile interval.  Both procedures use 10,000 resamples.  We use
$\widehat{\Delta}$ for the contrast observed on the evaluated paired cohort.

The AlfWorld interior endpoint maximum and HotPotQA endpoint reversal are
empirical findings on the evaluated grid.  The intervals quantify uncertainty
about the contrast magnitudes beyond these cases.  In particular, full replay
minus full withholding is $-1.0$ point on HotPotQA, with an interval that
includes zero.  These observations characterize schedule sensitivity in the
evaluated settings; they do not establish a generally optimal replay schedule
or causal harm from \Reflexion as a whole.

\subsection{Case-level heterogeneity}
\label{app:lf-replay-heterogeneity}

The schedules recover different subsets of the shared first-trial failures.
Among 58 such AlfWorld failures, the best single fixed schedule recovers 33,
whereas the union of all five schedules recovers 49.  The numbers recovered by
exactly $0,1,2,3,4,$ and $5$ schedules are $9,15,7,7,13,$ and $7$.  Among 64
HotPotQA failures, the best single schedule recovers 14 and the union recovers
18; the corresponding support counts are $46,2,3,5,4,$ and $4$.

For the three displayed schedules $\{0,.25,1\}$ on AlfWorld, 86/134 cases are
solved by all three; 5 only by $\varepsilon=.25$; 4 only by full replay; 6 only
by full withholding; 11 by replay and $\varepsilon=.25$; 7 by withholding and
$\varepsilon=.25$; 1 by the two endpoints; and 14 by none.  On HotPotQA,
42/100 are solved by all three, 2 by exactly one schedule, 8 by exactly two,
and 48 by none.

The complete per-case heatmap and recovery-support distribution appear in
\cref{fig:lf-replay-support}.

\subsection{Realized replay behavior}
\label{app:lf-replay-behavior}

For every realized later episode in the five-condition grid (with duplicate
named endpoints excluded), an exact repeat reproduces the complete normalized
task-action sequence of an earlier failed episode for the same case.  Exact
repeats recover in 0/304 AlfWorld episodes and 0/157 HotPotQA episodes.  Changed
sequences recover in 137/704 episodes (19.5\%) and 59/989 episodes (6.0\%),
respectively.  Every observed recovery therefore changes the failed sequence,
but most changed sequences still fail.

The randomized gates permit a descriptive exposure-conditioned calculation
over the three interior conditions.  For episode $j$, let $Z_j$ indicate
replay, $R_j$ indicate exact repetition, and
$\pi_j(1)=1-\varepsilon_j$, $\pi_j(0)=\varepsilon_j$.  The normalized
inverse-probability estimate is
\[
\widehat p_z=
\frac{\sum_j\mathbb{1}[Z_j=z]R_j/\pi_j(z)}
     {\sum_j\mathbb{1}[Z_j=z]/\pi_j(z)},\qquad z\in\{0,1\}.
\]
Replay versus withholding gives exact-repeat rates of 10.9\% versus 37.6\% on
AlfWorld and 4.9\% versus 20.0\% on HotPotQA.

Exact-action novelty is the fraction of action occurrences in an episode whose
lowercased, whitespace-normalized strings are absent from earlier episodes of
the same case and condition.  Novelty alone is not sufficient for recovery:
AlfWorld recovery rises from 8.4\% in the lowest novelty bin to
24.7\% in the highest, whereas HotPotQA peaks at 8.9\% in $[.25,.5)$ and falls
to 4.3\% in $[.5,.75)$.

The complete repetition, novelty, and recovery summaries appear in
\cref{fig:lf-replay-behavior}.

Taken together, the exposure surface, case overlap, and realized trajectories
show that retaining reflection and exposing it to the Actor are empirically
distinct under a finite trial budget.  The changing replay advantage and
complementary recoveries reveal an empirical EE trade-off between using failure
feedback and preserving alternative continuations.  We use this observation
to design \VEX{} and test its value in the primary evaluation.

%% file: appendices/scheduler.tex
\section{\VEX{} Prompt, Execution Semantics, and Process Cases}
\label{app:lf-dsst}

\subsection{\VEX{} Runtime Loop}
\label{app:lf-dsst-loop}

The released controller implements the loop in
\cref{alg:lf-dsst-loop}: the runtime resets and executes one complete Actor
trial, records only its
benchmark-visible executed Action--Observation trace after failure, makes one scheduler
call when another complete trial remains, and passes that response unchanged
to the next Actor.  The language model makes the joint exploration--exploitation
decision under the remaining trial budget without parameter updates.
The runtime neither parses the requested
sections nor executes a separate search algorithm over the textual tree.

\subsection{Complete input schema and requested output schema}
\label{app:lf-dsst-schema}

\begin{table}[H]
\centering
\small
\setlength{\tabcolsep}{4pt}
\renewcommand{\arraystretch}{1.1}
\begin{tabular*}{\textwidth}{@{\extracolsep{\fill}}
>{\raggedright\arraybackslash\ttfamily}p{0.37\textwidth}
>{\raggedright\arraybackslash}p{0.55\textwidth}@{}}
\toprule
Input field & Exact role and exclusion boundary \\
\midrule
task\_or\_task\_family\_scope & Stable goal or task-family scope.  It excludes
identifiers and values that must be rebound after reset. \\
basic\_environment\_rules & Public mechanics and action-interface limits; no
solution route, oracle annotation, or hidden state. \\
remaining\_complete\_attempts & Integer $R=T-t$ after the counted failed
attempt. \\
\makecell[l]{complete\_ordered\_failure\_\\only\_ao\_forest} & Every eligible failed attempt
in execution order, pairing the attempt index with the released
\texttt{canonical\_policy\_ao\_trace} field (the name is retained for artifact
compatibility).  It excludes successful trajectories,
unexecuted candidates, and model-invented outcomes. \\
\bottomrule
\end{tabular*}
\caption{Input fields passed to the \VEX{} scheduler and their information
boundaries.}
\label{tab:lf-app-dsst-input}
\end{table}

\begin{table}[H]
\centering
\small
\setlength{\tabcolsep}{4pt}
\renewcommand{\arraystretch}{1.1}
\begin{tabular*}{\textwidth}{@{\extracolsep{\fill}}
>{\raggedright\arraybackslash\ttfamily}p{0.22\textwidth}
>{\raggedright\arraybackslash}p{0.70\textwidth}@{}}
\toprule
Requested section & Model-visible schema \\
\midrule
EVIDENCE & Compact comparisons grounded by attempt and A--O references.
Issued actions are not evidence that they worked; the subsequent observation
or terminal feedback establishes the outcome. \\
TREE & Hierarchical outcome-relevant questions and end-to-end,
observation-conditioned policy leaves.  Every leaf has a short ID and exactly
one status: \texttt{[NEXT]}, \texttt{[RESERVE]}, or \texttt{[PARK]}. \\
ALLOCATION & One integer quota for every existing leaf ID.  Exactly one leaf
is \texttt{NEXT}; \texttt{NEXT} and retained \texttt{RESERVE} leaves receive
positive quotas; every \texttt{PARK} leaf receives zero; quotas sum exactly to
$R$. \\
NEXT-ATTEMPT POLICY & The selected direction, supported decisions to preserve,
decisions to change, reset-local information to rebind, any observed
reset-stable affordance needed under the within-attempt limit, and a visible
condition to check before completion. \\
\bottomrule
\end{tabular*}
\caption{Requested structure of a \VEX{} response.  The controller passes the
response to the Actor without runtime parsing or repair.}
\label{tab:lf-app-dsst-output}
\end{table}

\subsection{Prompt-defining decision semantics}
\label{app:lf-dsst-prompt}

\begin{table}[H]
\centering
\small
\setlength{\tabcolsep}{4pt}
\renewcommand{\arraystretch}{1.1}
\begin{tabular*}{\textwidth}{@{\extracolsep{\fill}}
>{\raggedright\arraybackslash}p{0.22\textwidth}
>{\raggedright\arraybackslash}p{0.70\textwidth}@{}}
\toprule
Semantic object & Prompt requirement \\
\midrule
Decision space & Nodes ask semantic questions whose alternatives could change
terminal outcome.  The tree is not an old action route, interface-operation
list, environment-state graph, or checklist. \\
Policy leaves & Each leaf specifies one complete observation-conditioned
policy direction; siblings differ in an outcome-relevant decision, preserve
all explicit task constraints, and fit the supplied within-attempt Actor
decision-block and A--O limits. \\
Evidence status & Actor-visible observations and terminal feedback are
evidence.  Untested branches and explanations from model prior knowledge
remain hypotheses.  A failed complete policy need not invalidate every
supported subdecision in its trajectory. \\
Reset boundary & Old reset-local identifiers and values cannot be reused.
They must be rebound from the next task and observation; only observed
reset-stable affordances may be carried forward. \\
Finite cross-trial horizon & Quotas distribute all $R$ remaining complete trials.  At
$R=1$, the scheduler must choose the strongest feasible success-directed
policy, not information-only exploration whose result cannot be used later. \\
Prohibited output & No executable action code, hidden task facts, unrebound
local values, or private chain-of-thought. \\
\bottomrule
\end{tabular*}
\caption{Decision semantics specified in the \VEX{} prompt.  Three synthetic
examples illustrate these semantics without contributing branches or answers
to the current instance.}
\label{tab:lf-app-dsst-semantics}
\end{table}

The allocation is advisory and is recomputed before each subsequent trial that
follows a counted failure.  The prompt designates the
\texttt{NEXT-ATTEMPT POLICY} as guidance for the immediately following counted
trial.  Because the complete response is handed off unparsed, the runtime does
not isolate that section from the other model-authored text; its ``selected''
status is a designation in the prompt, not a runtime-enforced constraint.  If the trial fails
and another trial remains, the scheduler is asked to rebuild statuses and
quotas from the current forest and smaller $R$; the runtime neither tracks
reservations nor repairs an invalid semantic plan through a free extra
generation.

These requirements specify the prompt's requested behavior, not fields
guaranteed by runtime validation.  In the realized \texttt{sample\_0009}
response, all four requested sections are present and the quotas sum to $R_1=5$.

\subsection{Primary scheduler generation configuration}
\label{app:lf-dsst-config}

All reported runs use the same three-shot \VEX{} prompt.  The primary MiniWoB
matrix uses episode indices 20--29.  WebShop evaluation uses 100 goals,
indices 0--99, from the official human-test split.

\begin{table}[H]
\centering
\small
\setlength{\tabcolsep}{4.8pt}
\begin{tabular*}{\textwidth}{@{\extracolsep{\fill}}llll@{}}
\toprule
Model profile & Temperature & Reasoning/thinking & Shared scheduler fields \\
\midrule
DeepSeek-V4-Flash & 0 where supported & Thinking disabled &
4,096-token cap; seed 42 \\
GPT-5.4 nano & Not sent & Effort \texttt{none}; minimized &
4,096-token cap; seed 42 \\
GLM-4.7-FlashX & 0 where supported & Thinking disabled &
4,096-token cap; seed 42 \\
\bottomrule
\end{tabular*}
\caption{Primary \VEX{} scheduler configuration.  The complete response is
retained; there is no semantic repair or free resampling.}
\label{tab:lf-app-dsst-config}
\end{table}

The complete prompt and its three synthetic examples are defined by
\texttt{DSST\_ICL\_PROMPT} in
\path{vex2_code_data_supplement/code/vex2/prompts.py} in the accompanying supplement.

\subsection{Realized \VEX{} Trajectory}
\label{app:lf-dsst-realized-record}

\Cref{fig:lf-dsst-information-path-main} visualizes GPT-5.4-nano WebShop
\texttt{sample\_0009}.  After a $T_1$ purchase with reward $0.6667$, the
scheduler allocates $\mathbf{a}_1=(3,2,0)$ across its policy alternatives.
The next Actor receives this response unchanged and succeeds at $T_2$.

\FloatBarrier

\subsection{Additional Paired \VEX{} Cases}
\label{app:lf-dsst-cases}

The two cases below show how \VEX{} and \Reflexion diverge after the same first
attempt in the primary GPT-5.4-nano WebShop setting.  Task, reset instance,
Actor, $T=6$ budget, and first-attempt trajectory are matched.  \VEX{} succeeds
earlier in 15 of the 100 paired cases; the table states the deterministic
selection rule for each example.  Setting-level conclusions use the complete
paired cohort.

\begin{table}[H]
\centering
\footnotesize
\setlength{\tabcolsep}{4pt}
\renewcommand{\arraystretch}{1.12}
\begin{tabular*}{\textwidth}{@{\extracolsep{\fill}}
>{\raggedright\arraybackslash}p{0.24\textwidth}
>{\raggedright\arraybackslash}p{0.68\textwidth}@{}}
\toprule
Field & Record \\
\midrule
Case and selection predicate &
\textbf{Sample 0017: preserve a verified option while changing identity.}
Lowest-index case satisfying these conditions in which \VEX{} preserves an observed
task-named option, changes product, and succeeds at $T_2$ while \Reflexion
remains unsolved. \\
Shared $T_1$ &
\texttt{B07VLKYJNH}; MN4; buy; terminal failure (reward 0.6667). \\
\Reflexion continuation &
$T_2$ opens \texttt{B07571SBH8} without MN4; $T_3$--$T_6$ repeat
\texttt{B07VLKYJNH} with MN4.  Unsolved at $T_6$. \\
\VEX{} allocation, continuation, and outcome &
$R=5$: different sensitive-skin product \texttt{NEXT}=3; same product with
evidence check \texttt{RESERVE}=2; exact repeat \texttt{PARK}=0.  At $T_2$,
open \texttt{B077PR9TL4}, select MN4, and buy; success. \\
\bottomrule
\end{tabular*}

\vspace{0.8\baselineskip}

\begin{tabular*}{\textwidth}{@{\extracolsep{\fill}}
>{\raggedright\arraybackslash}p{0.24\textwidth}
>{\raggedright\arraybackslash}p{0.68\textwidth}@{}}
\toprule
Field & Record \\
\midrule
Case and selection predicate &
\textbf{Sample 0057: sustain outcome-distinct alternatives.}
Lowest-index \VEX{}-only success satisfying these conditions and attaining five exact post-$T_1$ product repeats
by \Reflexion while \VEX{} opens at least three distinct products. \\
Shared $T_1$ &
\texttt{B088G1K5ZQ} kosher sea salt; buy; terminal failure (reward 0.5). \\
\Reflexion continuation &
$T_2$--$T_6$ reopen and buy \texttt{B088G1K5ZQ} five times.  Unsolved at
$T_6$. \\
\VEX{} allocation, continuation, and outcome &
$R=5$: alternative product \texttt{NEXT}=3; same product with option check
\texttt{RESERVE}=1; new-query alternative \texttt{RESERVE}=1; exact repeat
\texttt{PARK}=0.  Open \texttt{B004VEQUIW}, \texttt{B088G1K5ZQ}, and
\texttt{B08JNCQBGF} over $T_2$--$T_4$; success at $T_4$. \\
\bottomrule
\end{tabular*}
\caption{Two paired WebShop cases in which \VEX{} changes the continuation after
the shared first-trial failure.  Sample 0017 preserves a verified option while
changing product identity; Sample 0057 maintains outcome-distinct alternatives
while \Reflexion repeats one product.}
\label{tab:lf-app-dsst-cases}
\end{table}

%% file: appendices/reproducibility.tex
\FloatBarrier
\section{Reproducibility and Resource Accounting}
\label{app:lf-reproducibility}

\subsection{Primary execution profiles}
\label{app:lf-reproducibility-profiles}

Primary benchmark execution used CPU-only x86-64 hosts running Ubuntu 22.04.5
LTS or NixOS 25.05; model inference ran at the provider-hosted endpoints and
used no local GPU.  MiniWoB used CPython 3.12.10, Chromium 137.0.7151.68,
the matching ChromeDriver release, and Selenium 4.46.0.  WebShop used
CPython 3.10.20 and OpenJDK 11.0.31.  The analysis and diagnostic environments
use Python 3.12, and the release includes separate pinned dependency files for
analysis, diagnostics, MiniWoB, and WebShop.

\begin{table}[!ht]
\centering
\small
\setlength{\tabcolsep}{4pt}
\renewcommand{\arraystretch}{1.1}
\begin{tabular*}{\textwidth}{@{\extracolsep{\fill}}
>{\raggedright\arraybackslash}p{0.20\textwidth}
>{\raggedright\arraybackslash}p{0.25\textwidth}
>{\raggedright\arraybackslash}p{0.25\textwidth}
>{\raggedright\arraybackslash}p{0.22\textwidth}@{}}
\toprule
Profile & Provider interface & Deliberation setting & Transport and usage \\
\midrule
DeepSeek-V4-Flash & Official DeepSeek API via an OpenAI-compatible client &
Thinking disabled & Automatic transport retries disabled;
provider usage required and retained \\
GPT-5.4 nano & OpenAI API & Reasoning effort \texttt{none}; minimized
reasoning mode & Automatic transport retries disabled; provider usage required
and retained \\
GLM-4.7-FlashX & Zhipu BigModel API; model ID
\texttt{glm-4.7-flashx} & Thinking disabled & Automatic transport retries
disabled; provider usage required and retained \\
\bottomrule
\end{tabular*}
\caption{Primary provider profiles.  Provider-reported usage contains
prompt and completion tokens and is retained with the exported records.}
\label{tab:lf-app-reproducibility-profiles}
\end{table}

The primary configurations fix the cohorts as follows.  MiniWoB uses all 64
task families and episode indices
20--29, giving 10 reset-seed sequences per family, with the full per-episode
seed sequence paired across methods.  WebShop uses the fixed cohort of goal
indices 0--99 from the official 500-goal human-test split and resets each goal
to its entry state.  In MiniWoB, each successive trial uses a new reset seed
within the same task family.  Thus SR@6 measures success within a paired
sequence of up to six fresh instances, whereas WebShop measures success across
repeated attempts on the same goal.  Neither setting grants intermediate-state
restoration.

MiniWoB permits 128 dispatched A--O transitions per complete episode and a
task-specific number of Actor action blocks.  The block limit is two for
\texttt{book-flight}, \texttt{terminal}, and \texttt{use-autocomplete}; three
for \texttt{login-user} and \texttt{login-user-popup}; ten for
\texttt{guess-number} and \texttt{tic-tac-toe}; and one for the other 57
families.  Each block may contain multiple primitive actions, but another
observation-conditioned model decision requires another block.  Exhausting
either limit ends the trial.  These limits are identical across methods.
WebShop permits 15 A--O transitions per trial.  Both use $T=6$ as the single pre-specified
complete-trial budget rather than an estimated optimum, exact shared
memory-free $T_1$, early stopping upon success, and the same failure-eligibility
rule.  Every complete trial reduces the remaining trial budget.  Only failures
with a valid benchmark-visible record containing at least one dispatched A--O
pair are eligible: their benchmark-visible and method-internal records extend
$\mathcal F_t^{(m)}$ and $\mathcal Q_t^{(m)}$, respectively.  The shared $T_1$ includes its
benchmark-visible record, method-internal Actor record, and eligibility bit.
The \Reflexion visibility condition is specified in
\cref{sec:lf-vrlbench}; the implementation retains every eligible-failure
reflection and imposes no additional harness-side truncation.

The secondary GLM-5.1 MiniWoB setting retains the same 64 task families but
uses episode indices 10--19 rather than the primary profiles' 20--29.  Its
comparisons are paired within that secondary cohort; because model and cohort
change together, it is not a backbone-only comparison with the primary rows.

%% file: paper_arxiv.bbl
{\small

\begin{thebibliography}{23}
\providecommand{\natexlab}[1]{#1}
\providecommand{\url}[1]{\texttt{#1}}
\expandafter\ifx\csname urlstyle\endcsname\relax
  \providecommand{\doi}[1]{doi: #1}\else
  \providecommand{\doi}{doi: \begingroup \urlstyle{rm}\Url}\fi

\bibitem[Auer et~al.(2002)Auer, Cesa-Bianchi, and Fischer]{auer2002finite}
Peter Auer, Nicol{\`o} Cesa-Bianchi, and Paul Fischer.
\newblock Finite-time analysis of the multiarmed bandit problem.
\newblock \emph{Machine Learning}, 47\penalty0 (2--3):\penalty0 235--256, 2002.

\bibitem[{DeepSeek-AI}(2026)]{deepseekai2026v4}
{DeepSeek-AI}.
\newblock Models and pricing: {DeepSeek-V4-Flash}.
\newblock \url{https://api-docs.deepseek.com/quick_start/pricing/}, 2026.
\newblock Official API documentation, accessed 2026-08-10.

\bibitem[Gou et~al.(2024)Gou, Shao, Gong, Shen, Yang, Duan, and
  Chen]{gou2024critic}
Zhibin Gou, Zhihong Shao, Yeyun Gong, Yelong Shen, Yujiu Yang, Nan Duan, and
  Weizhu Chen.
\newblock {CRITIC}: Large language models can self-correct with
  tool-interactive critiquing.
\newblock In \emph{International Conference on Learning Representations}, 2024.

\bibitem[Hao et~al.(2023)Hao, Gu, Ma, Hong, Wang, Wang, and
  Hu]{hao2023reasoning}
Shibo Hao, Yi~Gu, Haodi Ma, Joshua~Jiahua Hong, Zhen Wang, Daisy~Zhe Wang, and
  Zhiting Hu.
\newblock Reasoning with language model is planning with world model.
\newblock In \emph{Proceedings of the 2023 Conference on Empirical Methods in
  Natural Language Processing}, pp.\  8154--8173, 2023.

\bibitem[Kapoor et~al.(2025)Kapoor, Stroebl, Siegel, Nadgir, and
  Narayanan]{kapoor2025agentsthatmatter}
Sayash Kapoor, Benedikt Stroebl, Zachary~S. Siegel, Nitya Nadgir, and Arvind
  Narayanan.
\newblock {AI} agents that matter.
\newblock \emph{Transactions on Machine Learning Research}, 2025.
\newblock URL \url{https://openreview.net/forum?id=Zy4uFzMviZ}.

\bibitem[Kapoor et~al.(2026)Kapoor, Stroebl, Kirgis, Nadgir, Siegel, Wei, Xue,
  Chen, Chen, Utpala, Ndzomga, Oruganty, Luskin, Liu, Yu, Arora, Hahm, Trivedi,
  Sun, Lee, Jin, Mai, Zhou, Zhu, Bommasani, Kang, Song, Henderson, Su, Liang,
  and Narayanan]{kapoor2026hal}
Sayash Kapoor, Benedikt Stroebl, Peter Kirgis, Nitya Nadgir, Zachary~S. Siegel,
  Boyi Wei, Tianci Xue, Ziru Chen, Felix Chen, Saiteja Utpala, Franck Ndzomga,
  Dheeraj Oruganty, Sophie Luskin, Kangheng Liu, Botao Yu, Amit Arora, Dongyoon
  Hahm, Harsh Trivedi, Huan Sun, Juyong Lee, Tengjun Jin, Yifan Mai, Yifei
  Zhou, Yuxuan Zhu, Rishi Bommasani, Daniel Kang, Dawn Song, Peter Henderson,
  Yu~Su, Percy Liang, and Arvind Narayanan.
\newblock Holistic agent leaderboard: The missing infrastructure for {AI} agent
  evaluation.
\newblock In \emph{The Fourteenth International Conference on Learning
  Representations}, 2026.
\newblock URL \url{https://openreview.net/forum?id=vUaY1t64ZZ}.

\bibitem[Liu et~al.(2018)Liu, Guu, Pasupat, Shi, and
  Liang]{liu2018reinforcement}
Evan~Zheran Liu, Kelvin Guu, Panupong Pasupat, Tianlin Shi, and Percy Liang.
\newblock Reinforcement learning on web interfaces using workflow-guided
  exploration.
\newblock In \emph{International Conference on Learning Representations}, 2018.

\bibitem[Madaan et~al.(2022)Madaan, Tandon, Clark, and
  Yang]{madaan2022memprompt}
Aman Madaan, Niket Tandon, Peter Clark, and Yiming Yang.
\newblock {MemPrompt}: Memory-assisted prompt editing with user feedback.
\newblock In \emph{Proceedings of the 2022 Conference on Empirical Methods in
  Natural Language Processing}, pp.\  2833--2861, 2022.

\bibitem[Madaan et~al.(2023)Madaan, Tandon, Gupta, Hallinan, Gao, Wiegreffe,
  Alon, Dziri, Prabhumoye, Yang, Gupta, Majumder, Hermann, Welleck,
  Yazdanbakhsh, and Clark]{madaan2023selfrefine}
Aman Madaan, Niket Tandon, Prakhar Gupta, Skyler Hallinan, Luyu Gao, Sarah
  Wiegreffe, Uri Alon, Nouha Dziri, Shrimai Prabhumoye, Yiming Yang, Shashank
  Gupta, Bodhisattwa~Prasad Majumder, Katherine Hermann, Sean Welleck, Amir
  Yazdanbakhsh, and Peter Clark.
\newblock Self-refine: Iterative refinement with self-feedback.
\newblock In \emph{Advances in Neural Information Processing Systems}, 2023.

\bibitem[Merrill et~al.(2026)Merrill, Shaw, Carlini, Li, Raj, Bercovich, Shi,
  Shin, Walshe, Buchanan, Shen, Ye, Lin, Poulos, Wang, Jitsev, Nezhurina, Lu,
  Mastromichalakis, Xu, Chen, Liu, Zhang, Chen, Kashyap, Uslu, Li, Wu, Yan,
  Bian, Sharma, Sun, Dillmann, Anand, Lanpouthakoun, Koopah, Hu, Guha, Dreiman,
  Zhu, Krauth, Zhong, Muennighoff, Amanfu, Tan, Pimpalgaonkar, Aggarwal, Lin,
  Lan, Zhao, Liang, Wang, Wang, Zhou, Heineman, Liu, Trivedi, Yang, Lin,
  Shetty, Yang, Omi, Raoof, Li, Zhuo, Lin, Dai, Wang, Chai, Zhou, Wahdany, She,
  Hu, Dong, Zhu, Cui, Saiyed, Kolbeinsson, Rytting, Marten, Wang, Dimakis,
  Konwinski, and Schmidt]{merrill2026terminalbench}
Mike~A Merrill, Alexander~Glenn Shaw, Nicholas Carlini, Boxuan Li, Harsh Raj,
  Ivan Bercovich, Lin Shi, Jeong~Yeon Shin, Thomas Walshe, E.~Kelly Buchanan,
  Junhong Shen, Guanghao Ye, Haowei Lin, Jason Poulos, Maoyu Wang, Jenia
  Jitsev, Marianna Nezhurina, Di~Lu, Orfeas~Menis Mastromichalakis, Zhiwei Xu,
  Zizhao Chen, Yue Liu, Robert Zhang, Leon~Liangyu Chen, Anurag Kashyap,
  Jan-Lucas Uslu, Jeffrey Li, Jianbo Wu, Minghao Yan, Song Bian, Vedang Sharma,
  Ke~Sun, Steven Dillmann, Akshay Anand, Andrew Lanpouthakoun, Bardia Koopah,
  Changran Hu, Etash~Kumar Guha, Gabriel H.~S. Dreiman, Jiacheng Zhu, Karl
  Krauth, Li~Zhong, Niklas Muennighoff, Robert~Kwesi Amanfu, Shangyin Tan,
  Shreyas Pimpalgaonkar, Tushar Aggarwal, Xiangning Lin, Xin Lan, Xuandong
  Zhao, Yiqing Liang, Yuanli Wang, Zilong Wang, Changzhi Zhou, David Heineman,
  Hange Liu, Harsh Trivedi, John Yang, Junhong Lin, Manish Shetty, Michael
  Yang, Nabil Omi, Negin Raoof, Shanda Li, Terry~Yue Zhuo, Wuwei Lin, Yiwei
  Dai, Yuxin Wang, Wenhao Chai, Shang Zhou, Dariush Wahdany, Ziyu She, Jiaming
  Hu, Zhikang Dong, Yuxuan Zhu, Sasha Cui, Ahson Saiyed, Arinbj{\"o}rn
  Kolbeinsson, Christopher~Michael Rytting, Ryan Marten, Yixin Wang, Alex
  Dimakis, Andy Konwinski, and Ludwig Schmidt.
\newblock Terminal-bench: Benchmarking agents on hard, realistic tasks in
  command line interfaces.
\newblock In \emph{The Fourteenth International Conference on Learning
  Representations}, 2026.
\newblock URL \url{https://openreview.net/forum?id=a7Qa4CcHak}.

\bibitem[{OpenAI}(2026)]{openai2026gpt54nano}
{OpenAI}.
\newblock {GPT-5.4 nano} model.
\newblock \url{https://developers.openai.com/api/docs/models/gpt-5.4-nano},
  2026.
\newblock Official model documentation, accessed 2026-08-10.

\bibitem[Shinn et~al.(2023)Shinn, Cassano, Gopinath, Narasimhan, and
  Yao]{shinn2023reflexion}
Noah Shinn, Federico Cassano, Ashwin Gopinath, Karthik Narasimhan, and Shunyu
  Yao.
\newblock Reflexion: Language agents with verbal reinforcement learning.
\newblock In \emph{Advances in Neural Information Processing Systems}, 2023.

\bibitem[Sutton \& Barto(2018)Sutton and Barto]{sutton2018reinforcement}
Richard~S. Sutton and Andrew~G. Barto.
\newblock \emph{Reinforcement Learning: An Introduction}.
\newblock MIT Press, 2 edition, 2018.

\bibitem[Suzgun et~al.(2026)Suzgun, Yuksekgonul, Bianchi, Jurafsky, and
  Zou]{suzgun2025dynamiccheatsheet}
Mirac Suzgun, Mert Yuksekgonul, Federico Bianchi, Dan Jurafsky, and James Zou.
\newblock Dynamic cheatsheet: Test-time learning with adaptive memory.
\newblock In \emph{Proceedings of the 19th Conference of the European Chapter
  of the Association for Computational Linguistics (Volume 1: Long Papers)},
  pp.\  7080--7106, 2026.
\newblock \doi{10.18653/v1/2026.eacl-long.333}.
\newblock URL \url{https://aclanthology.org/2026.eacl-long.333/}.

\bibitem[Wang et~al.(2025)Wang, Mao, Fried, and Neubig]{wang2024awm}
Zhiruo Wang, Jiayuan Mao, Daniel Fried, and Graham Neubig.
\newblock Agent workflow memory.
\newblock In \emph{Proceedings of the 42nd International Conference on Machine
  Learning}, volume 267 of \emph{Proceedings of Machine Learning Research},
  pp.\  63897--63911, 2025.
\newblock URL \url{https://proceedings.mlr.press/v267/wang25bx.html}.

\bibitem[Yao et~al.(2022)Yao, Chen, Yang, and Narasimhan]{yao2022webshop}
Shunyu Yao, Howard Chen, John Yang, and Karthik Narasimhan.
\newblock Webshop: Towards scalable real-world web interaction with grounded
  language agents.
\newblock In \emph{Advances in Neural Information Processing Systems}, 2022.

\bibitem[Yao et~al.(2023)Yao, Yu, Zhao, Shafran, Griffiths, Cao, and
  Narasimhan]{yao2024tree}
Shunyu Yao, Dian Yu, Jeffrey Zhao, Izhak Shafran, Thomas~L. Griffiths, Yuan
  Cao, and Karthik Narasimhan.
\newblock Tree of thoughts: Deliberate problem solving with large language
  models.
\newblock In \emph{Advances in Neural Information Processing Systems}, 2023.

\bibitem[{Z.AI}(2026)]{zai2026glm47}
{Z.AI}.
\newblock {GLM-4.7} developer documentation.
\newblock \url{https://docs.z.ai/guides/llm/glm-4.7}, 2026.
\newblock Official documentation for the GLM-4.7 series and the API model
  identifier \texttt{glm-4.7-flashx}, accessed 2026-08-10.

\bibitem[Zhang et~al.(2026)Zhang, Hu, Upasani, Ma, Hong, Kamanuru, Rainton, Wu,
  Ji, Li, Thakker, Zou, and Olukotun]{zhang2026ace}
Qizheng Zhang, Changran Hu, Shubhangi Upasani, Boyuan Ma, Fenglu Hong,
  Vamsidhar Kamanuru, Jay Rainton, Chen Wu, Mengmeng Ji, Hanchen Li, Urmish
  Thakker, James Zou, and Kunle Olukotun.
\newblock Agentic context engineering: Evolving contexts for self-improving
  language models.
\newblock In \emph{International Conference on Learning Representations}, 2026.
\newblock URL \url{https://openreview.net/forum?id=eC4ygDs02R}.

\bibitem[Zhang et~al.(2024)Zhang, Tang, Wu, Wang, Shen, Hou, Tan, Li, Zhuang,
  and Lu]{zhang2024agentpro}
Wenqi Zhang, Ke~Tang, Hai Wu, Mengna Wang, Yongliang Shen, Guiyang Hou, Zeqi
  Tan, Peng Li, Yueting Zhuang, and Weiming Lu.
\newblock Agent-pro: Learning to evolve via policy-level reflection and
  optimization.
\newblock In \emph{Proceedings of the 62nd Annual Meeting of the Association
  for Computational Linguistics (Volume 1: Long Papers)}, pp.\  5348--5375,
  2024.
\newblock \doi{10.18653/v1/2024.acl-long.292}.
\newblock URL \url{https://aclanthology.org/2024.acl-long.292/}.

\bibitem[Zhao et~al.(2024)Zhao, Huang, Xu, Lin, Liu, and Huang]{zhao2024expel}
Andrew Zhao, Daniel Huang, Quentin Xu, Matthieu Lin, Yong-Jin Liu, and Gao
  Huang.
\newblock Expel: {LLM} agents are experiential learners.
\newblock In \emph{Proceedings of the AAAI Conference on Artificial
  Intelligence}, 2024.
\newblock URL \url{https://ojs.aaai.org/index.php/AAAI/article/view/29936}.

\bibitem[Zheng et~al.(2024)Zheng, Wang, Wang, and An]{zheng2023synapse}
Longtao Zheng, Rundong Wang, Xinrun Wang, and Bo~An.
\newblock Synapse: Trajectory-as-exemplar prompting with memory for computer
  control.
\newblock In \emph{International Conference on Learning Representations}, 2024.
\newblock URL \url{https://openreview.net/forum?id=Pc8AU1aF5e}.

\bibitem[Zhou et~al.(2024)Zhou, Yan, Shlapentokh-Rothman, Wang, and
  Wang]{zhou2024lats}
Andy Zhou, Kai Yan, Michal Shlapentokh-Rothman, Haohan Wang, and Yu-Xiong Wang.
\newblock Language agent tree search unifies reasoning, acting, and planning in
  language models.
\newblock In \emph{Proceedings of the 41st International Conference on Machine
  Learning}, 2024.

\end{thebibliography}

}
